\documentclass[11pt,journal]{IEEEtran}

\usepackage{graphicx}
\usepackage[cmex10]{amsmath}

\usepackage{algorithmic}
\usepackage{array}

\usepackage{color}
\usepackage[usenames,dvipsnames]{xcolor}

\usepackage[tight,footnotesize]{subfigure}
\usepackage{afterpage}
\usepackage[ruled,commentsnumbered]{algorithm2e}
\usepackage[compress]{cite}
\usepackage{amssymb}
\usepackage{setspace}
\usepackage{color}
\usepackage{soul}
\usepackage{url}

\newcommand{\argmin}{\operatornamewithlimits{argmin}}
\newcommand\RR[1]{\mbox{I}\!\mbox{R}^{#1}}

\def\x{{\mathbf x}}
\def\a{{\mathbf a}}

\def\xi{{\x_i}}

\def\y{{\mathbf y}}
\def\z{{\mathbf z}}
\def\U{{\mathbf U}}
\def\G{{\mathbf G}}

\def\Supp{{\mathcal{S}}}

\def\D{{\mathbf D}}

\def\Fi{{\boldsymbol \Phi}}

\def\W{{\mathbf W}}
\def\A{{\mathbf A}}

\def\T{{\mathcal T}}

\def\R{{\mathbf R}}
\def\P{{\mathbf P}}
\def\O{{\mathcal O}}

\def\X{{\mathbf X}}
\def\Y{{\mathbf Y}}
\def\P{{\mathbf P}}

\def\f{\mathbf{f}}
\def\g{{\mathbf g}}

\graphicspath{{figures/}}

\begin{document}

\title{Trainlets: Dictionary Learning in High Dimensions}

\author{Jeremias~Sulam,~\IEEEmembership{Student Member,~IEEE,}
        Boaz~Ophir,~
        Michael~Zibulevsky,~
        and~Michael~Elad,~\IEEEmembership{Fellow,~IEEE}

\thanks{Copyright (c) 2015 IEEE. Personal use of this material is permitted. However, permission to use this material for any other purposes must be obtained from the IEEE by sending a request to pubs-permissions@ieee.org.
The research leading to these results has received funding from the European Research Council under European Union’s Seventh Framework Programme, ERC Grant agreement no. 320649. This research was supported in part by the grant of Intel Collaborative Research Institute for Computational Intelligence (ICRI-CI), and by the Israel Science Foundation (ISF) grant number 1770/14.
 All authors are with the Computer Science Department, Technion Israel Institute of Technology.}

}

\maketitle

\IEEEpeerreviewmaketitle

\begin{abstract}
Sparse representations has shown to be a very powerful model for real world signals, and has enabled the development of applications with notable performance. Combined with the ability to learn a dictionary from signal examples, sparsity-inspired algorithms are often achieving state-of-the-art results in a wide variety of tasks. Yet, these methods have traditionally been restricted to small dimensions mainly due to the computational constraints that the dictionary learning problem entails. In the context of image processing, this implies handling small image patches.

In this work we show how to efficiently handle bigger dimensions and go beyond the small patches in sparsity-based signal and image processing methods. We build our approach based on a new cropped wavelet decomposition, which enables a multi-scale analysis with virtually no border effects. We then employ this as the base dictionary within a double sparsity model to enable the training of adaptive dictionaries. To cope with the increase of training data, while at the same time improving the training performance, we present an Online Sparse Dictionary Learning (OSDL) algorithm to train this model effectively, enabling it to handle millions of examples. This work shows that dictionary learning can be up-scaled to tackle a new level of signal dimensions, obtaining large adaptable atoms that we call \emph{trainlets}.

\end{abstract}


\vspace{-0.3cm}
\section{Introduction}
\label{Sect:Intro}
Sparse representations over redundant dictionaries have shown to be a very powerful model for many real world signals, enabling the development of applications with notable performance in many signal and image processing tasks \cite{Elad_Book}.
The basic assumption of this model is that natural signals can be expressed as a sparse linear combination of atoms, chosen from a collection called a \emph{dictionary}.
Formally, for a signal $\y\in \RR{n}$, this can be described by $\y = \D\x$, where
$\D \in \RR{n\times m}, (n<m)$ is a redundant dictionary that contains the atoms as its columns, and $\x\in \RR{m}$ is the representation vector.

Given the signal $\y$, finding its representation can be done in terms of the following sparse approximation problem:
\begin{equation}\label{Eq:Sparse Coding}
\min_{\x}\|\x\|_{0} \quad \mbox{subject to}
\quad \|\mathbf{y}-{\bf Dx}\|_{2}\leq\epsilon,
\end{equation}
where $\epsilon$ is a permitted deviation in the representation accuracy, and the expression $\|\x\|_0$ is a count of the number of non-zeroes in the vector $\x$. The process of solving the above optimization problem is commonly referred to as sparse-coding.
Solving this problem is in general NP-hard, but several greedy algorithms and other relaxations
methods allow us to solve the problem exactly under certain conditions \cite{Bruckstein2009} and obtain useful approximate solutions
in more general settings. These methods include MP \cite{Mallat1993}, OMP \cite{Pati1993a}, BP \cite{Chen2001} and FOCUSS \cite{Gorodnitsky1997} among others.

A fundamental element in this problem is the choice of the dictionary $\D$. While some analytically-defined dictionaries (or transformations) such as the overcomplete Discrete Cosine Transform (ODCT) or Wavelet dictionaries were used originally, learning the dictionary from signal examples  for a specific task has shown to perform significantly better \cite{Rubinstein2010_dict}. This adaptivity to the data allows sparsity-inspired algorithms to achieve state-of-the-art results in many tasks. The dictionary learning problem can be written as:
\begin{equation}\label{Eq:Dict Learning}
\argmin_{\D,\X}\frac{1}{2}\|\Y-\D\X\|_{F}^{2} \quad \mbox{subject
to} \quad \|\x_{i} \|_{0}\leq p \quad \forall i,
\end{equation}
where $\Y \in \RR{n\times N}$ is a matrix containing N signal examples, and $\X \in \RR{m\times N}$ are the corresponding sparse vectors, both
ordered column wise. Several iterative methods have been proposed to handle this task \cite{Aharon2006,Engan1999,Mairal2010}.
Due to the computational complexity of this problem, all these methods have been restricted to relatively small signals. When dealing with high-dimensional data, the common approach is to partition the signal into small blocks, where the dictionary learning problem is more feasible.

In the context of image processing, small signals imply handling small image patches. Most state-of-the-art methods for image restoration exploit such a localized patch based approach \cite{Dabov2007,Mairal2009,Zoran2011}. In this setting, small overlapping patches ($7\times7$ - $11\times11$) are extracted from the corrupted
image and treated relatively independently according to some image model \cite{Dabov2006,Zoran2011}, sparse representations being a popular choice \cite{Dong2011,Yang2010,Elad2006,Romano2014}.
The full image estimation is then formed by merging together the small restored patches by overlapping and averaging.

Some works have attempted to handle larger two dimensional patches (i.e., greater than $16\times 16$) with some success. In \cite{Ophir2011}, and later in \cite{Sulam2014}, traditional K-SVD is applied in the Wavelet domain. These works implicitly manage larger patches while keeping the atom dimension small, noting that small patches of Wavelet coefficients translate to large regions in the image domain. In the context of Convolutional Networks, on the other hand, the work in \cite{Burger2012} has reported encouraging state-of-art result on patches of size $17\times 17$.

Though adaptable, explicit dictionaries are computationally expensive to apply. Some efforts have been done in designing fast dictionaries that can be both applied and learned efficiently. This requirement implies constraining the degrees of freedom of the explicit matrix in some way, i.e. imposing some structure on the dictionary.
One such possibility is the search for adaptable separable dictionaries, as in \cite{Hawe2013}, or the search of a dictionary which is an image in itself as in \cite{Aharon2008,Benoit2011}, lowering the degrees of freedom and obtaining (close to) shift invariant atoms.

Another, more flexible alternative, has been the pursuit of sparse dictionaries \cite{Rubinstein2010,Yaghoobi09}. In these works the dictionary is composed of a multiplication of two matrices, one of which is sparse. The work in \cite{Lemagoarou15} takes this idea a step further, composing a dictionary from the multiplication of a sequence of sparse matrices.
In the interesting work reported in \cite{Chabiron2013} the dictionary is modeled as a collection of convolutions with sparse kernels, lowering the complexity of the problem and enabling the approximation of popular analytically-defined atoms.
All of these works, however, have not addressed dictionary learning on real data of considerably higher dimensions or with a considerably large dataset.

A related but different model from the one posed in Equation \eqref{Eq:Dict Learning} is the analysis model \cite{Elad2007,Rubinstein2014}. In this framework, a dictionary $\W$ is learned such that $\|\W \y\|_0 \ll n $. A close variant is the Transform Learning model, where it is assumed that  $\W\y \approx \x $ and $\|\x\|_0 \ll n $, as presented in \cite{Ravishankar2013a}. This framework presents interesting advantages due to the very cheap sparse coding stage (a thresholding operation). 
An online transform learning approach was presented in \cite{Ravishankar2015}, and a sparse transform model was presented in \cite{Ravishankar2013}, enabling the training on bigger image patches. In our work, however, we constrain ourselves to the study of synthesis dictionary models.

We give careful attention to the model proposed in \cite{Rubinstein2010}. In this work a double sparse model is proposed by combining a fixed separable dictionary with an adaptable sparse component. This lowers the degrees of freedom of the problem in Equation \eqref{Eq:Dict Learning}, and provides a feasible way of treating high dimensional signals. However, the work reported in \cite{Rubinstein2010} concentrated on 2D and 3D-DCT as a base-dictionary, thus restricting its applicability to relatively small patches.

In this work we expand on this  model, showing how to efficiently handle bigger dimensions and go beyond the small patches in sparsity-based signal and image processing methods. This model provides the flexibility of incorporating multi-scale properties in the learned dictionary, a property we deem vital for representing larger signals. For this purpose, we propose to replace the fixed base-dictionary with a new multi-scale one.
We build our approach on \emph{cropped} wavelets, a multi-scale decomposition which overcomes the limitations of the traditional wavelet transform to efficiently represent small images (expressed often in the form of severe border effects).

Another aspect that has limited the training of large dictionaries has been the amount of data required and the corresponding amount of computations involved. As the signal size increases, a (significant) increase in the number of training examples is needed in order to effectively learn the inherent data structure. While traditional dictionary learning algorithms require many sweeps of the whole training corpus, this is no longer feasible in our context.
Instead, we look to online learning methods, such as Stochastic Gradient Decent (SGD) \cite{bottou98}. These methods have gained prominence in recent years with the advent of big data, and have been used in the context of traditional (unstructured) dictionary learning \cite{Mairal2010} and in training the special structure of the Image Signature Dictionary \cite{Aharon2008}.
We present an Online Sparse Dictionary Learning (OSDL) algorithm to effectively train the double-sparsity model. This approach allows us to handle very large training sets while using high dimensional signals, achieving faster convergence than the batch alternative and providing a better treatment of local minima, which are abundant in non-convex dictionary learning problems.


To summarize, this paper introduces a novel online dictionary learning algorithm,
which builds a structured dictionary based on the double-sparsity format. The base-dictionary proposed is a fully-separable cropped Wavelets that has virtually no boundary effects. The overall dictionary learning algorithm can be trained on a corpus of millions of examples, and is capable of representing images of size $64\times 64$ and even more, while keeping the training, the memory, and the computational load reasonable and manageable. This high-dimensional dictionary learning framework, termed \emph{trainlets}, shows that global dictionaries for entire images are feasible and trainable. We demonstrate the applicability of the proposed algorithm and its various ingredients in this paper, and we accompany this work with a freely available software package.

This paper is organized as follows.
In section \ref{Sect:SparseDicts} we review sparse dictionary models.
In section \ref{Sect:CrWave} we introduce the Cropped Wavelets and show their advantages over standard Wavelets.
In section \ref{Sect:OSDL} we present the Online Sparse Dictionary Learning algorithm, comparing it to the alternative method for training such a model, Sparse KSVD, and to the Online Dictionary Learning algorithm of \cite{Mairal2010}, which trains an unconstrained (dense) dictionary.
In section  \ref{Sect:Exp} we present results from several experiments and applications to image processing, demonstrating the benefits of our proposed method, and in section \ref{Sect:Summary} we conclude the paper.



\section{Sparse Dictionaries}
\label{Sect:SparseDicts}
Learning dictionaries for large signals requires adding some constraint to the dictionary, otherwise signal diversity and the number of training examples needed make the problem intractable. Often, these constraints are given in terms of a certain structure.
One such approach is the double-sparsity model \cite{Rubinstein2010}. In this model the dictionary is assumed to be a multiplication of a fixed operator $\Fi$ (we will refer to it as the base dictionary) by a sparse adaptable matrix $\A$. Every atom in the effective dictionary $\D$ is therefore a linear combination of few and arbitrary atoms from the base dictionary. Formally, this means that the training procedure requires solving the following problem:
\begin{equation}
\min_{\A,\X}\ \frac{1}{2}|| \Y - \Fi \A \X ||^2_F \quad \text{s.t.} \quad \left\{
	\begin{array}{ll}
		||\x_i||_{0} \leq p  & \forall i \\
		||\a_j||_{0} = k  & \forall j
	\end{array}.
\right.
\label{Eq:Sparse KSVD}
\end{equation}
Note that the number of columns in $\Fi$ and $\A$ might differ, allowing flexibility in the redundancy of the effective dictionary.
The authors in \cite{Rubinstein2010} used an over-complete Discrete Cosine Transform (ODCT) as the base dictionary in their experiments. Using Wavelets was proposed but never implemented due both to implementation issues (the traditional Wavelet transform is not entirely separable) and to the significant border-effects Wavelets have in small-to-medium sized patches. We address both of these issues in the following section.

As for the training of such a model, the update of the dictionary is now constrained by the number of non-zeros in the columns of $\A$.
In \cite{Rubinstein2010} a variant of the K-SVD algorithm (termed Sparse K-SVD) was proposed for updating the dictionary. As the work in \cite{Aharon2006}, this is a batch method that updates every atom sequentially. In the context of the double-sparsity structure, this task is converted into a sparse-coding problem, and approximated by the greedy OMP algorithm.

In the recent inspiring work reported in \cite{Lemagoarou15} the authors extended the double-sparsity model to a scenario where the base dictionary itself is a multiplication of several sparse matrices, that are to be learned.
While this structure allows for a clear decrease in the computational cost of applying the dictionary, its capacity to treat medium-size problems is not explored.
The proposed algorithm involves a hierarchy of matrix factorizations with multiple parameters to be set, such as the number of levels and the sparsity of each level. 

\section{A New Wavelets Dictionary}
\label{Sect:CrWave}

The double sparsity model relies on a base-dictionary which should be computationally efficient to apply.
The ODCT dictionary has been used for this purpose in \cite{Rubinstein2010}, but its applicability to larger signal sizes is weak. Indeed, as the patch size grows -- getting closer to an image size -- the more desirable a multi-scale analysis framework becomes\footnote{It is well known that when working with small patches in an image, a transform such as the 2D-DCT is highly effective. This is the reason for the success of DCT in JPEG. When the patch grows to become a small image, DCT is in fact highly ineffective as it insists of periodicity all over the support of the image. It is then Wavelets and its variants that emerge as an appealing alternative. Again, this explains the migration to Wavelets and frames when it comes to JPEG-2000 and global image restoration methods.}. The separability of the base dictionary provides a further decrease in the computational complexity. Applying two (or more) 1D dictionaries on each dimension separately is typically much more efficient than an equivalent non-separable multi-dimensional dictionary. We will combine these two characteristics as guidelines in the design of the base dictionary for our model. 	

\subsection{Optimal Extensions and Cropped Wavelets}
\label{Sec:CropedWave}

\begin{figure*}
\begin{center}
\includegraphics[trim = 120 10 120 10, width = 0.9\textwidth]{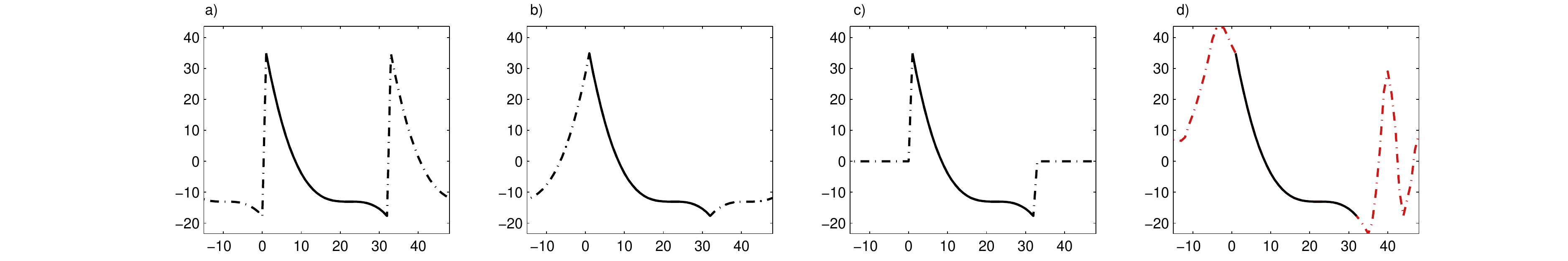}
\caption{Different border treatments: a) periodic, b) symmetric, c) zero-padding, and d) the resulting optimized extension signal $\bar{\f} = \W_s\g_w$.
}
\label{fig:Border_extension}
\end{center}
\vspace{-0.6cm}
\end{figure*}

The two dimensional Wavelet transform has shown to be very effective in sparsifying natural (normal sized) images. When used to analyze small or medium sized images, not only is the number of possible decomposition scales limited, but more importantly the border effects become a serious limitation. Other works have pointed out the importance of the boundary conditions in the context of deconvolution \cite{Almeida2013,Reeves2005}. However, our approach is different from these, as we will focus on the basis elements rather than on the signal boundaries, and in the pursuit of the corresponding coefficients.

In order to build (bi-)orthogonal Wavelets over a finite (and small) interval, one usually assumes their periodic or symmetric extension onto an infinite axis. A third alternative, zero-padding, assumes the signal is zero outside of the interval. However, none of these alternatives provides an optimal approximation of the signal borders. In general, all these methods do not preserve their vanishing moments at the boundary of the interval, leading to additional non-zero coefficients corresponding to the basis functions that overlap with the boundaries \cite{Mallat_AwaveletTour}. An alternative is to modify the Wavelet filters such that they preserve their vanishing moments at the borders of the interval, although constructing such Wavelets while preserving their orthogonality is complicated \cite{Cohen1993}.

We begin our derivation by looking closely at the zero-padding case. Let $\mathbf{f} \in \R^n$ be a finite signal.
Consider $\bar{\f} = \P \f$, the zero-padded version of $\f$, where $\P \in \R^{L\times n}$, $L>n$ ($L$ is ``big enough'').
Considering the Wavelet analysis matrix $\W_a$ of size $L\times L$, the Wavelet representation coefficients are obtained by applying the Discrete Wavelet Transform (DWT) to $\bar{\f}$, which can be written as $\g_w = \W_a \bar{\f}$. Note that this is just a projection of the (zero-padded) signal onto the orthogonal Wavelet atoms.

As for the inverse transform, the padded signal is recovered by applying the inverse Wavelet transform or Wavelet synthesis operator $\W_s$ ($\W_s=\W_a^{T}$, assuming orthogonal Wavelets), of size $L\times L$ to the coefficients $\g_w$. Immediately after, the padding is discarded (multiplying by $\P^T$) to obtain the final signal in the original finite interval:
\begin{equation}
\hat{\f} = \P^T \W_s \g_w = \P^T \W_s \left( \W_a \P \f\right) = \f.
\end{equation}

Zero-padding is not an option of preference because it introduces discontinuities in the function $\bar{\f}$ that result in large (and many) Wavelet coefficients, even if $\f$ is smooth inside the finite interval. This phenomenon can be understood from the following perspective: we are seeking the representation vector $\g_w$ that will satisfy the perfect reconstruction of $\f$,
\begin{equation}
\P^T \W_s \g_w = \f.
\end{equation}
The matrix $\P^T\W_s$ serves here as the effective dictionary that multiplies the representation in order to recover the signal. This relation is an under-determined linear system of equations with $n$ equations and $L$ unknowns, and thus it has infinitely many possible solutions.

In fact, zero padding chooses a very specific solution to the above system, namely, $\g_w = \W_a \P \f$. This is nothing but the projection of the signal onto the adjoint of the above-mentioned dictionary, since $\W_a \P = (\P^T \W_s)^T$. While this is indeed a feasible solution, such a solution is expected to have many non-zeros if the atoms are strongly correlated. This indeed occurs for the finite-support Wavelet atoms that intersect the borders, and which are cropped by $\P^T$.

To overcome this problem, we propose the following alternative optimization objective:
\begin{equation}
	\g_w = \underset{\g}{\arg\min} \| \g \|_0 \quad \text{s.t.}  \quad  \P^T\W_s \g  = \f,
	\label{CropeedProblem}
\end{equation}
i.e., seeking the sparsest solution to this under-determined linear system. Note that in performing this pursuit, we are implicitly extending the signal $\f$ to become $\bar{\f} = \W_s \g_w$, which is the smoothest possible with respect to the Wavelet atoms (i.e., it is sparse under the Wavelet transform). At the same time, we keep using the original Wavelet atoms with all their properties, including their vanishing moments. On the other hand, we pay the price of performing a pursuit instead of a simple back-projection. In particular, we use OMP to approximate the solution to this sparse coding problem. To conclude, our treatment of the boundary issue is obtained by applying the cropped Wavelets dictionary $\W_c = \P^T\W_s$, and seeking the sparsest representation with respect to it, implicitly obtaining an extension of $\f$ without boundary problems.

To illustrate our approach, in Fig. \ref{fig:Border_extension} we show the typical periodic, symmetric and zero-padding border extensions applied to a random smooth function, as well as the ones obtained by our method. As can be seen, this extension -- which is nothing else than Wavelet atoms that \emph{fit} in the borders in a natural way -- guarantees not to create discontinuities which result in denser representations\footnote{A similar approach was presented in \cite{Zhao_2000} in the context of compression. The authors proposed to optimally extend the borders of an irregular shape in the sense of minimal $l_1$-norm of the representation coefficients under a DCT transform.}. Note that we will not be interested in the actual extensions explicitly in our work.

\begin{figure}
\begin{center}
\includegraphics[trim = 40 20 40 20, width = 0.35\textwidth]{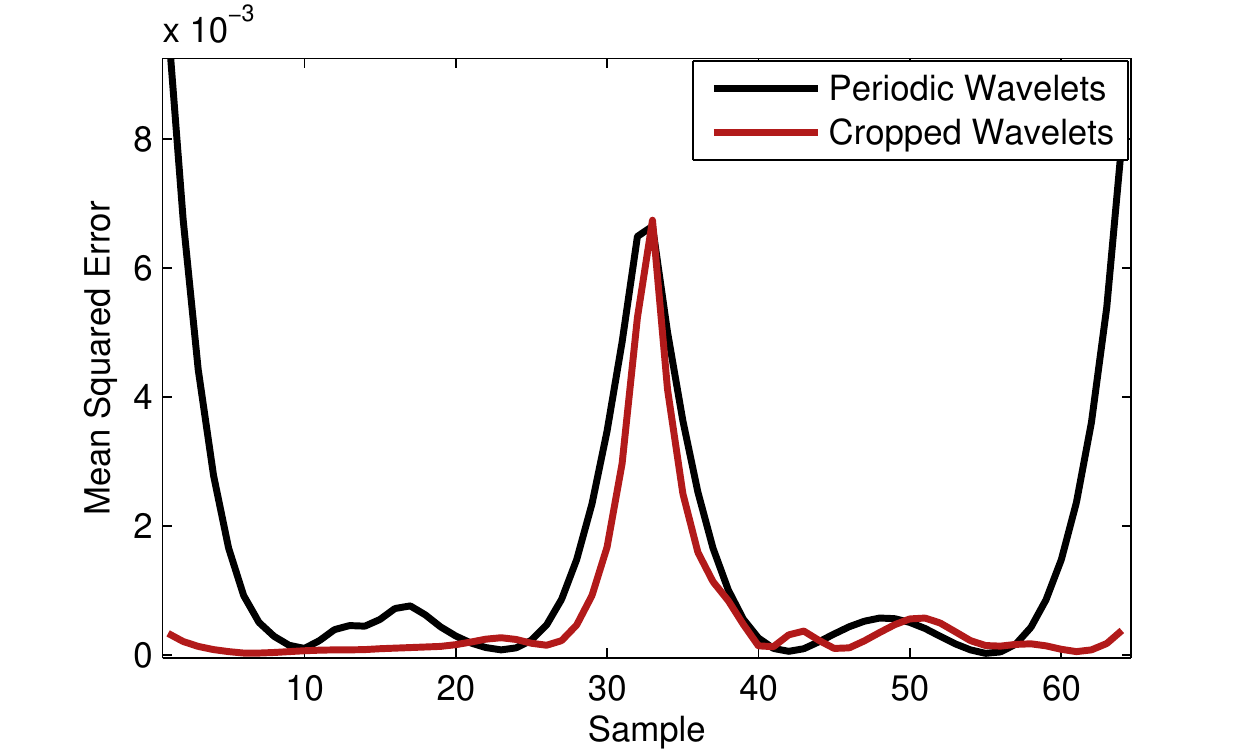}
\caption{Mean approximation (using 5 coefficients) error \emph{per sample} of smooth functions of length 64 with a discontinuity at sample 32. }
\label{fig:BorderStudy}
\end{center}
\vspace{-0.5cm}
\end{figure}
To provide further evidence on the better treatment of the borders by the cropped Wavelets, we present the following experiment. We construct 1,000 random smooth functions $\f$ of length 64 (3rd degree polynomials), and introduce a random step discontinuity at sample 32. These signals are then normalized to have unit $l_2$-norm. We approximate these functions with only 5 Wavelet coefficients\footnote{The m-term approximation with Wavelets is performed with the traditional non-linear approximation scheme. In this framework, orthogonal Wavelets with periodic extensions perform better than symmetric extensions or zero-padding, which we therefore omit from the comparison. We used for this experiment Daubechies Wavelets with 13 taps. All random variables were chosen from Gaussian distributions.},
and measure the energy of the point-wise (per sample) error (in $l_2$-sense) of the reconstruction. Fig.\ref{fig:BorderStudy} shows the mean distribution of these errors. As expected, the discontinuity at the center introduces a considerable error. However, the traditional (periodic) Wavelets also exhibit substantial errors at the borders. The proposed cropped Wavelets, on the other hand, manage to reduce these errors by avoiding the creation of extra discontinuities.

Practically speaking, the proposed cropped Wavelet dictionary can be constructed by taking a Wavelet synthesis matrix for signals of length $L$ and cropping it. Also, and because we will be making use of greedy pursuit methods, each atom is normalized to have unit $l_2$ norm. This way, the cropped Wavelets dictionary can be expressed as
\begin{equation}\nonumber
\Fi^c_1 = \mathbf{P}^T\ \W_s\ \boldsymbol{\mathcal{W}},
\end{equation}
where $\boldsymbol{\mathcal{W}}$ is a diagonal matrix of size $L\times L$ with values such that each atom (column) in $\Fi^c_1$ (of size $n\times L$) has a unit norm\footnote{Because the atoms in $\W_s$ are compactly supported, some of them may be identically zero in the central $n$ samples. These are discarded in the construction of $\Fi^c_1$.}.
The resulting transform is no longer orthogonal, but this -- now redundant -- Wavelet dictionary solves the borders issues of traditional Wavelets enabling for a lower approximation error.

Just as in the case of zero-padding, the redundancy obtained depends on the dimension of the signal, the number of decomposition scales and the length of the support of the Wavelet filters (refer to \cite{Mallat_AwaveletTour} for a thorough discussion). In practice, we set $L = 2^{\lceil log_2(n)\rceil +1}$; i.e, twice the closest higher power of 2 (which reduces to $L = 2n$ if $n$ is a power of two, yielding a redundancy of at most 2) guaranteeing a sufficient extension of the borders.

\subsection{A Separable 2-D Extension}

The one-dimensional Wavelet transform is traditionally extended to treat two-dimensional signals by constructing two-dimensional atoms as the separable product of two one-dimensional ones, per scale \cite{Mallat_AwaveletTour}. This yields three two-dimensional Wavelet functions at each scale $j$, implying a decomposition which is only separable per scale.
In practice, this means cascading this two-dimensional transform on the approximation band at every scale.

An alternative extension is a completely separable construction. Considering all the basis elements of the 1-D DWT (in \emph{all} scales) arranged column-wise in the matrix $\Fi_{1}$, the 2-D separable transform can be represented as the Kronecker product $\Fi_{2} = \Fi_{1} \otimes \Fi_{1}$. This way, all properties of the transform $\Fi_{1}$ translate to each of the dimensions of the 2-dimensional signal on which $\Fi_{2}$ is applied. Now, instead of cascading down a two-dimensional decomposition, the same 1-D Wavelet transform is applied first to all the columns of the image and then to all the rows of the result (or vice versa). In relatively small images, this alternative is simpler and faster to apply compared to the traditional cascade.
This modification is not only applicable to the traditional Wavelet transform, but also to the cropped Wavelets dictionary introduced above. In this 2-D set-up, both vertical and horizontal borders are implicitly extended to provide a sparser Wavelet representation.

\begin{figure}
\centering
\includegraphics[trim = 60 40 60 20, width = .38\textwidth]{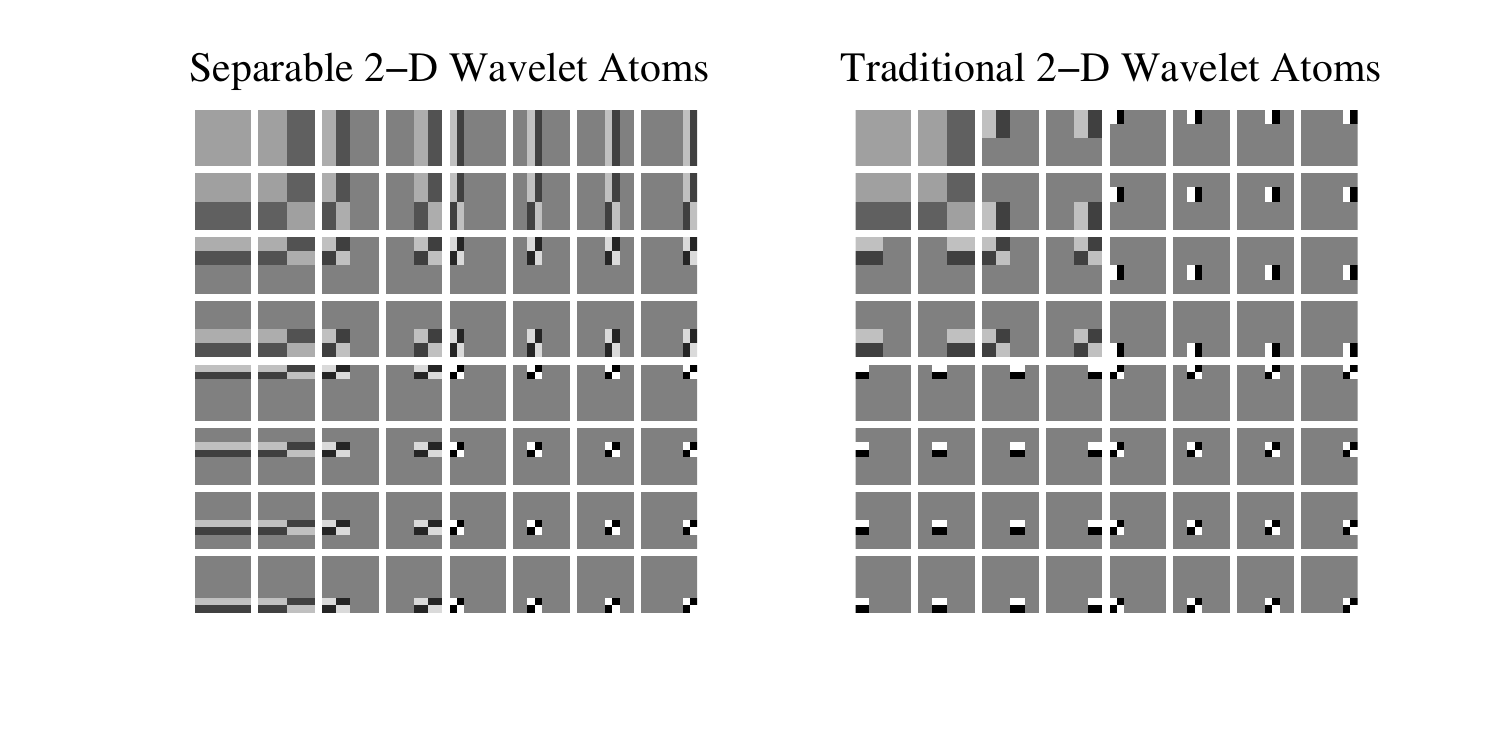}\\
\caption{2-D atoms of the Wavelet (Haar) transform for patches of size $8 \times 8$  -- the separable versus the the traditional construction.
}
\label{fig:SeparableDict}
\vspace{-0.4cm}
\end{figure}

We present in Fig. \ref{fig:SeparableDict} the 2-D atoms of the Wavelet (Haar) Transform for signals of size  $8\times 8$ as an illustrative example.
The atoms corresponding to the coarsest decomposition scale and the diagonal bands are the same in both separable and non-separable constructions. The difference appears in the vertical and horizontal bands (at the second scale and below). In the separable case we see elongated atoms, mixing a low scale in one direction with high scale in the other.

\subsection{Approximation of Real World Signals}

\begin{figure} \centering
\begin{tabular}{cc}
\includegraphics[trim = 190 200 10 160, width = .123\textwidth]{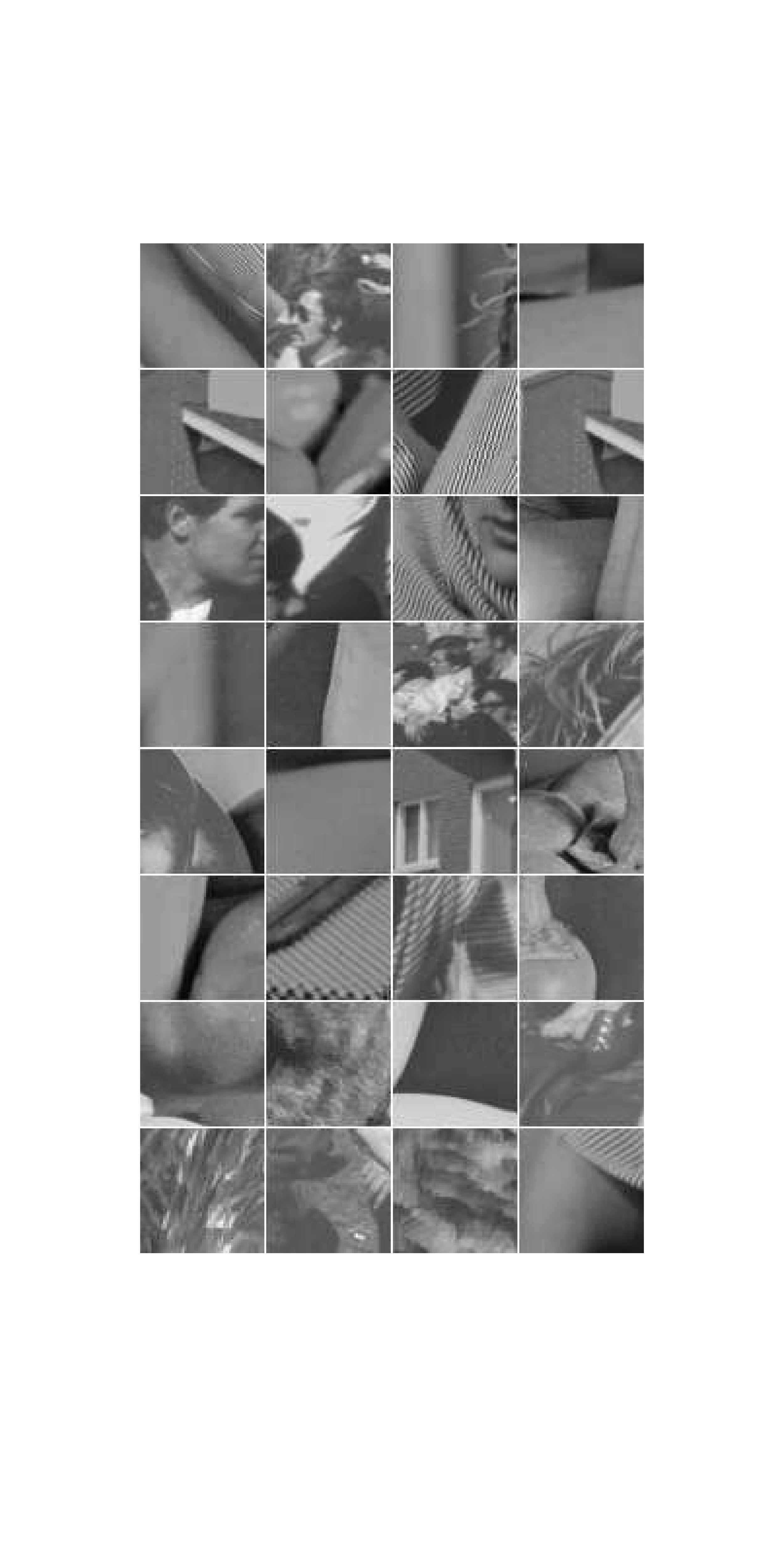} &
\includegraphics[trim = 5 10 60 80, width = .22\textwidth]{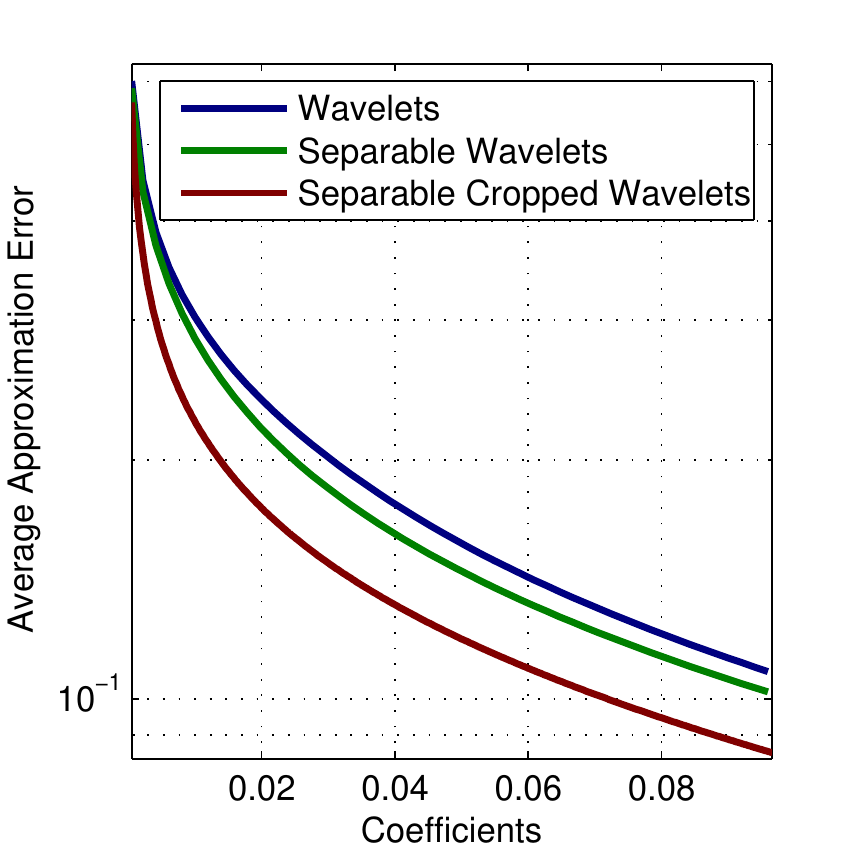} \\
\end{tabular}
\caption{Left: A random set of some of the images used to evaluate the M-Term approximation of the regular and the separable cropped Wavelets. Right: M-Term approximation by the traditional 2-D Wavelets and the separable and cropped Wavelets on real images of size $64\times64$.}
\label{fig:CrWaveFig}
\vspace{-0.5cm}
\end{figure}

While it is hard to rank the performance of separable versus non-separable \emph{analytical} dictionaries or transforms in the general case, we have observed that the separable Wavelet transform provides sparser representations than the traditional 2-D decomposition on small-medium size images. To demonstrate this, we take 1,000 image patches of size $64\times64$ from popular test images, and compare the m-term approximation achieved by the regular two-dimensional Wavelet transform, the completely separable Wavelet transform and our separable and cropped Wavelets.
A small subset of these patches is presented on the left of Fig. \ref{fig:CrWaveFig}. These large patches are in themselves small images, exhibiting the complex structures characteristic of real world images.

As we see from the results in Fig. \ref{fig:CrWaveFig} (right), the separability provides some advantage over regular Wavelets in representing the image patches. Furthermore, the proposed separable cropped Wavelets give an even better approximation of the data with fewer coefficients.

Before concluding this section, we make the following remark. It is well known that Wavelets (separable or not) are far from providing an optimal representation for general images \cite{Mallat_AwaveletTour,CandesCurvelets,do2005contourlet}. Nonetheless, in this work these basis functions will be used only as the base dictionary, while our learned dictionary will consist of linear combinations thereof. It is up to the learning process to close the gap between the sub-optimal representation capability of the Wavelets, and the need for a better two dimensional representation that takes into account edge orientation, scale invariance, and more.

\vspace{-0.25cm}
\section{Online Sparse Dictionary Learning}
\label{Sect:OSDL}

As seen previously, the \emph{de-facto} method for training the doubly sparse model has been a batch-like process. When working with higher dimensional data, however, the required amount of training examples and the corresponding computational load increase. In this \emph{big-data} (or \emph{medium-data}) scenario, it is often unfeasible or undesired to perform several sweeps over the entire data set. In some cases, the dimensionality and the amount of data might restrict the learning process to only a couple of iterations. In this regime of work it may be impossible to even store all training samples in memory during the training process. In an extreme online learning set-up, each data sample is seen only once as new data flows in.

These reasons lead naturally to the formulation of an online training method for the double-sparsity model. In this section, we first introduce a dictionary learning method based on the Normalized Iterative Hard-Thresholding algorithm \cite{Blumensath2010}. We then use these ideas to propose an Online Sparse Dictionary Learning (OSDL) algorithm based on the popular Stochastic Gradient Descent technique, and show how it can be applied efficiently to our specific dictionary learning problem.

\subsection{NIHT-based Dictionary Learning}

A popular practice in dictionary learning, which has been shown to be quite effective, is to employ a block coordinate minimization over this non-convex problem. This often reduces to alternating between a sparse coding stage, throughout which the dictionary is held constant, and a dictionary update stage in which the sparse coefficients (or their support) are kept fixed.
We shall focus on the second stage, as the first remains unchanged, essentially applying sparse coding to a group of examples.
Embarking from the objective as given in Equation \eqref{Eq:Sparse KSVD}, the problem to consider in the dictionary update stage is the following:
\begin{equation}
\min_{\A} \underbrace{\frac{1}{2}|| \Y - \Fi \A \X ||^2_F}_{f(\A)} \quad \text{s.t.} \quad ||\a_j||_{0} = k \quad \forall j,
\label{Eq:DictUpdate}
\end{equation}
where $\Fi$ is the base dictionary of size $n\times L$ and $\A$ is a matrix of size $L\times m$ with $k$ non-zeros per column. Many dictionary learning methods undertake a sequential update of the atoms in the dictionary (\cite{Aharon2006,Mairal2010,Rubinstein2010}). Following this approach, we can consider $m$ minimization problems of the following form:
\begin{equation}
\min_{\a_j} \underbrace{\frac{1}{2}|| \mathbf{E}_j - \Fi \a_j \x^T_j ||^2_F}_{f(\a_j)} \quad \text{s.t.} \quad ||\a_j||_{0} = k,
\label{Eq:DictUpdate_atomwise}
\end{equation}
where $\mathbf{E}_j$ is the error given by $\Y-\sum_{i\neq j}\Fi \a_i \x^T_i$ and $\x^T_i$ denotes the $i$-th row of $\X$. This problem produces the $j$-th column in $\A$, and thus we sweep through \mbox{$j=1,\dots,m$} to update all of $\A$.

The Normalized Iterative Hard-Thresholding (NIHT) \cite{Blumensath2010} algorithm is a popular sparse coding method in the context of Compressed Sensing \cite{Blumensath2008}. This method can be understood as a projected gradient descent algorithm. We can propose a dictionary update based on the same concept. Note that we could rewrite the cost function in Equation \eqref{Eq:DictUpdate_atomwise} as $f(\a_j) = \frac{1}{2}|| \mathbf{E}_j - \mathcal{H}_j \a_j ||^2_F$, for an appropriate operator $\mathcal{H}_j$.
Written in this way, we can perform the dictionary update in terms of the NIHT by iterating:
\begin{equation}
\a_j^{t+1} = \mathcal{P}_k \left[  \a_j^{t} - \eta_j^t \ \mathcal{H}_j^* \left( \mathbf{E}_j - \mathcal{H}_j \a^t_j \right) \right],
\label{Eq:NIHT}
\end{equation}
where $\mathcal{H}_j^*$ is the adjoint of $\mathcal{H}_j$, $\mathcal{P}_k$ is a Hard-Thresholding operator that keeps the $k$ largest non-zeros (in absolute value), and $\eta_j^t$ is an appropriate step-size. Note that this algorithm implies iterating over Equation \eqref{Eq:NIHT} until convergence \emph{per} atom in the dictionary update stage.

The choice of the step size is critical. Noting that $\mathcal{H}_j^* \left( \Y - \mathcal{H}_j \a_j \right) = \nabla f(\a_j)$, in \cite{Blumensath2010} the authors propose to set this parameter per iteration as:
\begin{equation}
\eta_j^t = \frac{\|\nabla f(\a_j^t)_{S_j} \|_F^2}{\|\mathcal{H}\nabla f(\a_j^t)_{S_j} \|_F^2},
\label{Eq:StepSize1}
\end{equation}
where ${S_j}$ denotes the support of $\a_j^t$. With this step size, the estimate $\hat{\a}^{t+1}$ is obtained by performing a gradient step and hard-thresholding as in Equation \eqref{Eq:StepSize1}. Note that if the support of $\hat{\a}_j^{t+1}$ and $\a_j^{t}$ are the same, setting $\eta_j^t$ as in Equation \eqref{Eq:StepSize1} is indeed optimal, as it is the minimizer of the quadratic cost w.r.t. $\eta_j^t$.  In this case, we simply set $\a_j^{t+1} = \hat{\a}_j^{t+1}$. If the support changes after applying $\mathcal{P}_k$, however, the step-size must be diminished until a condition is met, guaranteeing a decrease in the cost function\footnote{The step size is decreased by $\eta_j^t = c\ \eta_j^t$, where $c<1$. We refer the reader to \cite{Blumensath2008} and \cite{Blumensath2010} for further details.}.
Following this procedure, the work reported in \cite{Blumensath2010} shows that the algorithm in Equation \eqref{Eq:NIHT} is guaranteed to converge to a local minimum of the problem in \eqref{Eq:DictUpdate_atomwise}.

Consider now the algorithm given by iterating between 1) sparse coding of all examples in $\Y$, and 2) atom-wise dictionary update with NIHT in Equation \eqref{Eq:DictUpdate_atomwise}. An important question that arises is: will this simple algorithm converge? Let us assume that the pursuit succeeds, obtaining the sparsest solution for a given sparse dictionary $\A$, which can indeed be guaranteed under certain conditions. Moreover, pursuit methods like OMP, Basis Pursuit and FOCUSS perform very well in practice when $k \ll n$ (refer to \cite{Bruckstein2009} for a thorough review).  For the cases where the theoretical guarantees are not met, we can adopt an external interference approach by comparing the best solution using the support obtained in the previous iteration to the one proposed by the new iteration of the algorithm, and choosing the best one. This small modification guarantees a decrease in the cost function at every sparse coding step.
The atom-wise update of the dictionary is also guaranteed to converge to a local minimum for the above mentioned choice of step sizes. Performing a series of these alternating minimization steps ensures a monotonic reduction in the original cost function in Equation \eqref{Eq:Dict Learning}, which is also bounded from below,  and thus convergence to a fixed point is guaranteed.

\subsection{From Batch to Online Learning}

As noted in \cite{Aharon2008,Mairal2010}, it is not compulsory to accumulate all the examples to perform an update in the gradient direction. Instead, we turn to a stochastic (projected) gradient descent approach. In this scheme, instead of computing the expected value of the gradient by the sample mean over all examples, we estimate this gradient over a single randomly chosen example $\y_i$. We then update the atoms of the dictionary based on this estimation using:
\begin{equation}
\a_j^{t+1} = \mathcal{P}_k \left[  \a_j^{t} - \eta^t \ \nabla f\left(\a_j^t,\y_i,\x_i,\right) \right].
\label{Eq:Dictupdate_AtomWise_Stochastic}
\end{equation}

Since these updates might be computationally costly (and because we are only performing an alternating minimization over problem \eqref{Eq:Sparse KSVD}), we might stop after a few iterations of applying Equation \eqref{Eq:Dictupdate_AtomWise_Stochastic}. We also restrict this update to those atoms that are used by the current example (since others have no contribution in the corresponding gradient). In addition, instead of employing the step size suggested by the NIHT algorithm, we employ the common approach of using decreasing step sizes throughout the iterations, which has been shown beneficial in stochastic optimization \cite{Bottou2012}. To this end, and denoting by $\eta_j^*$ the step size resulting from the NIHT, we employ an effective learning rate of $\frac{\eta_j^*}{1+t/T}$, with a manually set parameter $T$.
This modification does not compromise the guarantees of a decrease in the cost function (for the given random sample $i$), since this factor is always smaller than one.
We outline the basic stages of this method in Algorithm \ref{BasicAlgo}.
\begin{algorithm}
\setstretch{1.3}
\KwData{Training samples $\{\y_i\}$, base-dictionary $\Fi$, initial sparse matrix $\A^0$}
 \For{$i = 1,\dots,Iter$}{
 	Draw $\y_i$ at random\;
	$\x_i \leftarrow$ Sparse Code ($\y_i,\Fi,\A^i$)\;
	$\Supp_i = Support(\x_i)$\;
	\For{$j = 1,\dots,|\Supp_i|$}{
		
			Update $\a^{i+1}_{\Supp(j)}$  with Equation \eqref{Eq:Dictupdate_AtomWise_Stochastic} and step size $\frac{\eta_{\Supp(j)}^*}{1+i/T}$\;
	}
  }
\KwResult{Sparse Dictionary $\A$ }
\vspace{0.2cm}
\caption{Stochastic NIHT for Sparse Dictionary Learning.}
\label{BasicAlgo}
\end{algorithm}

An important question that now arises is whether shifting from a batch training approach to this online algorithm preserves the convergence guarantees described above. Though plenty is known in the field of stochastic approximations, most of the existing results address convergence guarantees for convex functions, and little is known in this area regarding projected gradient algorithms \cite{Bottou2008}. For non-convex cases, convergence guarantees still demand the cost function to be differentiable with continuous derivatives \cite{Bottou1998}. In our case, the $l_0$ pseudo-norm makes a proof of convergence challenging, since the problem becomes not only non-convex but also (highly) discontinuous.

That said, one could re-formulate the dictionary learning problem using a non-convex but continuous and differentiable penalty function\footnote{One of many such possibilities is $h(\a) = \sum_i \left(1-e^{-\alpha a_i^2}\right)$, replacing $\|\a\|_0$.}, moving from a constrained optimization problem to an unconstrained one. 
We conjecture that convergence to a fixed point of this problem can be reached under the mild conditions described in \cite{Bottou1998}.
Despite these theoretical benefits, we choose to maintain our initial formulation in terms of the $l_0$ measure for the sake of simplicity (note that we need no parameters other than the target sparsity). Practically,
we saw in all our experiments that convergence is reached, providing numerical evidence for the behavior of our algorithm.

\subsection{OSDL In Practice}

We now turn to describe a variant of the method described in Algorithm \ref{BasicAlgo}, and outline other implementation details.
The atom-wise update of the dictionary, while providing a specific step-size, is computationally slower than a global update. In addition, guaranteeing a decreasing step in the cost function implies a line-search per atom that is costly. For this reason we propose to replace this stage by a global dictionary update of the form
\begin{equation}
\A^{t+1} = \mathcal{P}_k \left[  \A^{t} - \eta^t \ \nabla f(\A^t) \right],
\label{Eq:GlobalUpdate}
\end{equation}
where the thresholding operator now operates in each column of its argument. While we could maintain a NIHT approach in the choice of the step-size in this case as well, we choose to employ
\begin{equation}
\eta^{\star} = \frac{|| \nabla f(\A_S)||_F}{||\Fi \nabla f(\A_S) X||_F}.
\label{eq:Eta}
\end{equation}
Note that this is the square-root of the value in Equation \eqref{Eq:StepSize1} and it may appear as counter-intuitive. We shall present a numerical justification of this choice in the following section.

Secondly, instead of considering a single sample $\y_t$ per iteration, a common practice in stochastic gradient descent algorithms is to consider mini-batches $\{\y_i\}$ of $N$ examples arranged in the matrix $\Y_t$.
As explained in detail in \cite{Rubinstein2008}, the computational cost of the OMP algorithm can be reduced by precomputing (and storing) the Gram matrix of the dictionary $\D$, given by $\G = \D^T\D$. In a regular online learning scheme, this would be infeasible due to the need to recompute this matrix for each example. In our case, however, the matrix needs only to be updated once per mini-batch. Furthermore, only a few atoms get updated each time. We exploit this by updating only the respective rows and columns of the matrix $\G$. Moreover, this update can be done efficiently due to the sparsity of the dictionary $\A$.

\begin{algorithm}[t]
\setstretch{1.3}
 \KwData{Training samples $\{\y_i\}$, base-dictionary $\Fi$, initial sparse matrix $\A^0$}
 Initialization: $\G_{\Fi} = \Fi^T\Fi$; $\U = \mathbf{0}$ \;
\For{$t = 1,\dots,T$}{
 	Draw a mini-batch $\Y_t$ at random\;
	$\X_t \leftarrow$ Sparse Code ($\Y_t,\Fi,\A^t,\G^t$)\;
	$\eta^t = ||\nabla f(\A_{\Supp}^t)||_F / \| \Fi \nabla f(\A_{\Supp}^t) \X^{\Supp}_t\|_F$\;
	$\U_{\Supp}^{t+1} = \gamma \U_{\Supp}^t + \eta^t \ \nabla f(\A_{\Supp}^t)$\;	
	$\A_{\Supp}^{t+1} = \mathcal{P}_k \left[  \A_{\Supp}^{t} - \U_{\Supp}^{t+1} \right]
$\;
	Update columns and rows of $\G$ by $\left(\A^{t+1}\right)^T \G_{\Fi} \A_{\Supp}^{t+1}$	
  }
\KwResult{Sparse Dictionary $\A$ }
\vspace{-0.0cm}
\caption{Online Sparse Dictionary Learning (OSDL) algorithm.}
\label{OSDL_Algorithm}
\end{algorithm}

Stochastic algorithms often introduce different strategies to regularize the learning process and try to avoid local minimum traps. In our case, we incorporate in our algorithm a momentum term $\U^t$ 
controlled by a parameter $\gamma \in \left[0,1\right]$. This term helps to attenuate oscillations and can speed up the convergence by incorporating information from the previous gradients. This algorithm, termed Online Sparse Dictionary Learning (OSDL) is depicted in Algorithm \ref{OSDL_Algorithm}.
In addition, many dictionary learning algorithms \cite{Aharon2006,Mairal2010} include the replacement of (almost) unused atoms and the pruning of similar atoms. We incorporate these strategies here as well, checking for such cases once every few iterations.

\subsection{Complexity Analysis}

We now turn to address the computational cost of the proposed online learning scheme. As was thoroughly discussed in \cite{Rubinstein2010}, the sparse dictionary enables an efficient sparse coding step. In particular, any multiplication by $\D$, or its transpose, has a complexity of $\mathcal{T}_D = \O( k m + \mathcal{T}_{\Fi})$, where $m$ is the number of atoms in $\Fi$ (assume for simplicity $\A$ square), $k$ is the atom sparsity and $\mathcal{T}_{\Fi}$ is the complexity of applying the base dictionary. For the separable case, this reduces to $\mathcal{T}_{\Fi} = \mathcal{O}(n\sqrt{m})$.

Using a sparse dictionary, the sparse coding stage with OMP (in its Cholesky implementation) is $\mathcal{O}( p n \sqrt{m} + p k m )$ per example. Considering $N$ examples in a mini-batch, and assuming $n \propto m$ and $p \propto n$, we obtain a complexity of $\mathcal{O}\left(N n^2 (\sqrt{m}+p) \right)$.

Moving to the update stage in the OSDL algorithm\footnote{We analyze the complexity of just the OSDL for simplicity. The analysis of Algorithm \ref{BasicAlgo} is similar, adding the complexity of the line search of the step sizes.}, calculating the gradient $\nabla f(A_{\Supp})$ has a complexity of $\T_{\nabla f} = \mathcal{O}((k |\Supp| + n \sqrt{m})N)$, and so does the calculation of the step size. Recall that $\Supp$ is the set of atoms used by the current samples, and that \mbox{$|\Supp| < m$}; i.e., the update is applied only on a subset of all the atoms. Updating the momentum variable grows as $\mathcal{O}(|\Supp|m)$, and the hard thresholding operator is $\mathcal{O}(|\Supp|m log(m))$. In a pessimistic approach, assume $|\Supp|\propto n$.

Putting these elements together, the OSDL algorithm has a complexity of $\mathcal{O}(N n^2(\sqrt{m}+k) + m^2 log(m))$ per mini-batch. The first term depends on the number of examples per mini-batch, and the second one depends only on the size of the dictionary. For high dimensions (large $n$), the first term is the leading one. Clearly, the number of non-zeros per atom $k$ determines the computational complexity of our algorithm. While in this study we do not address the optimal way of scaling $k$, experiments shown hereafter suggest that its dependency with $n$ might in fact be less than linear.
The sparse dictionary provides a computational advantage over the online learning methods using explicit dictionaries, such as \cite{Mairal2010}, which have complexity of $\mathcal{O}(N n^3)$.

\vspace{-0.25cm}
\section{Experiments}
\label{Sect:Exp}
\begin{figure*}
\centering
\includegraphics[trim = 120 50 100 30, width = .95\textwidth]{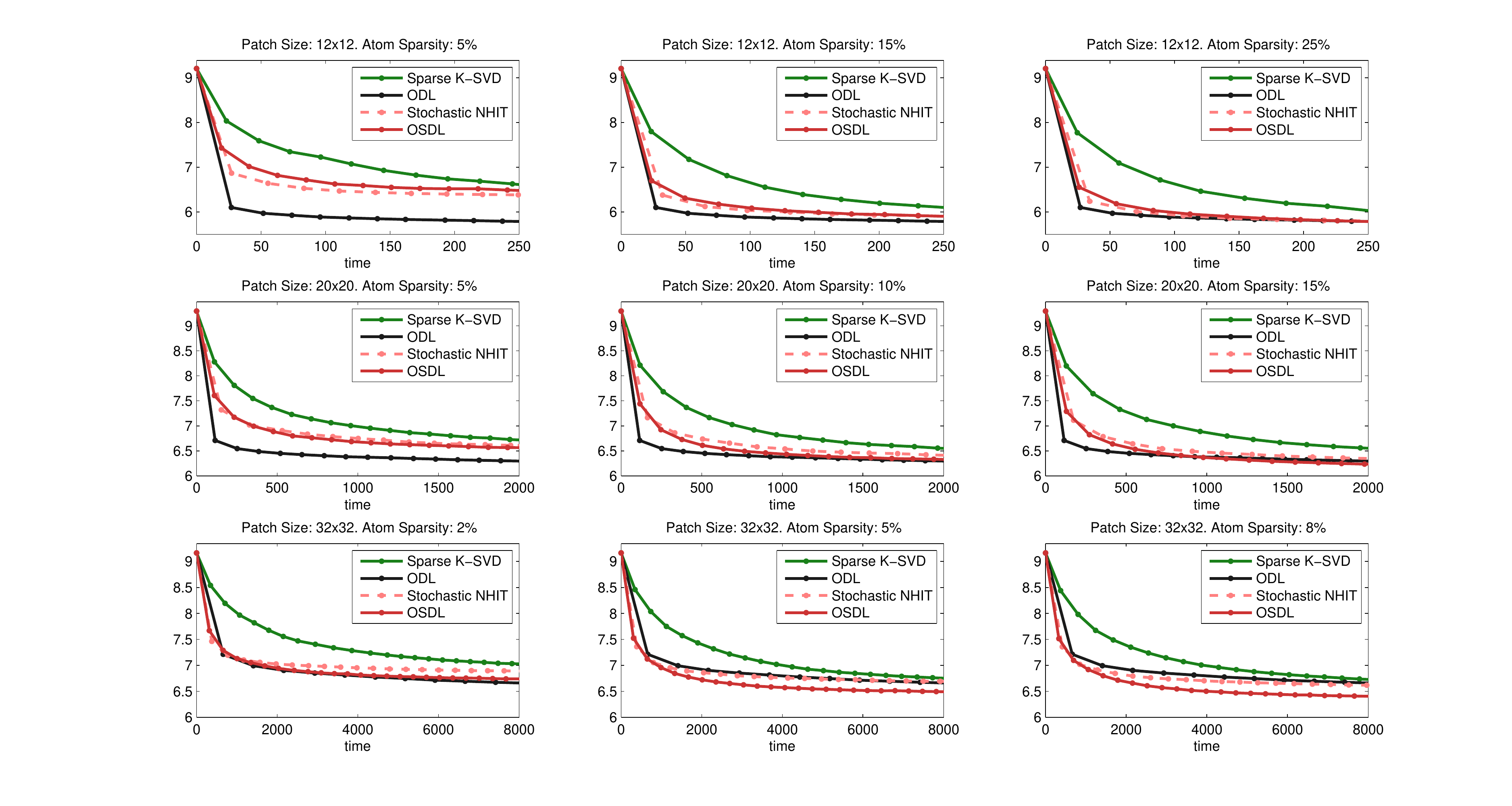}
\caption{Experiment 1: Dictionary learning by Sparse K-SVD, by the Stochastic NIHT presented in Algorithm \ref{BasicAlgo}, the ODL algorithm \cite{Mairal2009a} and by the Online Sparse Dictionary Learning (OSDL).}
\label{fig:Convergence1}
\vspace{-0.2cm}
\end{figure*}

In this section we present a number of experiments to illustrate the behaviour of the method presented in the previous section. We start with a detailed experiment on learning an image-specific dictionary. We then move on to demonstrations on image denoising and image compression. Finally we tackle the training of universal dictionaries on millions of examples in high dimensions.

\subsection{Image-Specific Dictionary Learning}

To test the behaviour of the proposed approach, we present the following experiment. We train an adaptive sparse dictionary in three setups of increasing dimension:  with patches of size $12\times12$, $20\times 20$ and $32\times 32$, all extracted from the popular image Lena, using a fixed number of non-zeros in the sparse coding stage (4, 10 and 20 non-zeros, respectively). We also repeat this experiment for different levels of sparsity of the dictionary $\A$. We employ the OSDL algorithm, as well as the method presented in Algorithm \ref{BasicAlgo} (in its mini-batch version, for comparison). We also include the results by Sparse K-SVD, which is the classical (batch) method for the double sparsity model, and the popular Online Dictionary Learning (ODL) algorithm \cite{Mairal2009a}. Note that this last method is an online method that trains a dense (full) dictionary. Training is done on 200,000 examples, leaving 30,000 as a test set.

The sparse dictionaries use the cropped Wavelets as their operator $\Fi$, built using the Symlet Wavelet with 8-taps. The redundancy of this base dictionary is 1.75 (in 1-D), and the matrix $\A$ is set to be square, resulting in a total redundancy of just over 3. For a fair comparison, we initialize the ODL method with the same cropped Wavelets dictionary. All methods use OMP in the sparse coding stage. Also, note that the ODL\footnote{We used the publicly available SPArse Modeling Software package, at http://spams-devel.gforge.inria.fr/.} algorithm is implemented entirely in C, while in our case this is only true for the sparse coding, giving the ODL somewhat of an advantage in run-time.

The results are presented in Fig. \ref{fig:Convergence1}, showing the representation error on the test set, where each marker corresponds to an epoch. The atom sparsity refers to the number of non-zeros per column of $\A$ with respect to the signal dimension (i.e., $5\%$ in the $12\times12$ case implies 7 non-zeros). Several conclusions can be drawn from these results. First, as expected, the online approaches provide a much faster convergence than the batch alternative. For the low dimensional case, there is little difference between Algorithm \ref{BasicAlgo} and the OSDL, though this difference becomes more prominent as the dimension increases. In these cases, not only does Algorithm \ref{BasicAlgo} converge slower but it also seems to be more prone to local minima.

As the number of non-zeros per atom grows, the representation power of our sparse dictionary increases. In particular, OSDL achieves the same performance as ODL for an atom sparsity of $25\%$ for a signal dimension of 144. Interestingly, OSDL and ODL achieve the same performance for decreasing number of non-zeros in $\A$ as the dimension increases: $10\%$ for the $20\times 20$ case and $\approx 2\%$ for the $32\times 32$. In this higher dimensional setting, not only does the sparse dictionary provide faster convergence but it also achieves a lower minimum. The lower degrees of freedom of the sparse dictionary prove beneficial in this context, where the amount of training data is limited and perhaps insufficient to train a full dictionary\footnote{Note that this limitation needed to be imposed for a comparison with Sparse K-SVD. Further along this section we will present a comparison without this limitation.}. This example suggests that indeed $k$ could grow slower than linearly with the dimension $n$.

\begin{figure}
\centering
\includegraphics[trim = 80 20 80 30, width = .3\textwidth]{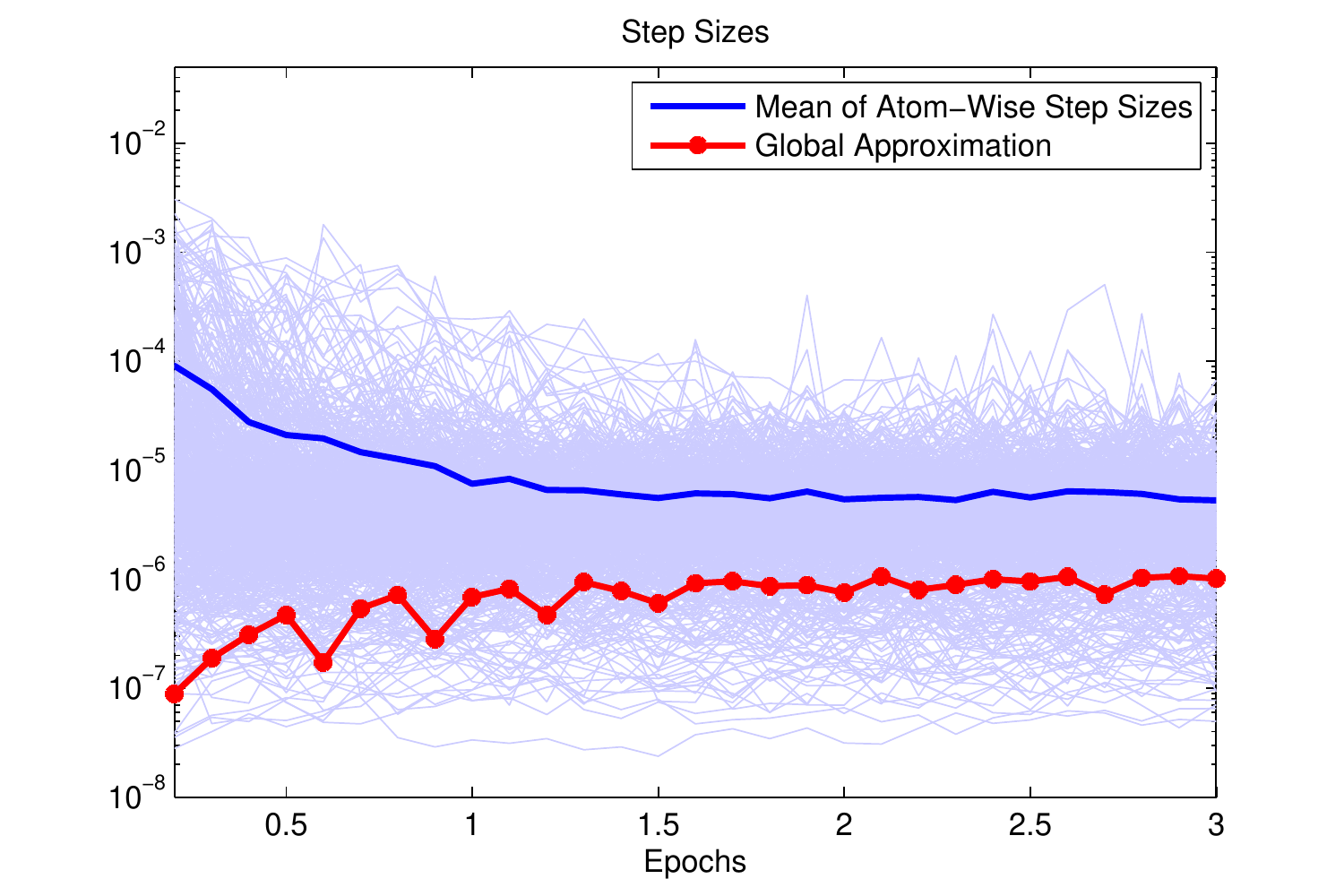}
\caption{Step sizes $\eta^*_j$ obtained by the atom-wise NIHT algorithm together with their mean value, and the global approximation by OSDL.}
\label{fig:StepSizes}
\vspace{-0.5cm}
\end{figure}

Before moving on, we want to provide some empirical evidence to support the choice of the step size in the OSDL algorithm. In Fig. \ref{fig:StepSizes} we plot the atom-wise step sizes obtained by Algorithm \ref{BasicAlgo}, $\eta^*_j$ (i.e., the optimal values from the NIHT perspective), together with their mean value, as a function of the iterations for the $12\times 12$ case for illustration. In addition, we show the global step sizes of OSDL as in Equation \eqref{eq:Eta}. As can be seen, this choice provides a fair approximation to the mean of the individual step sizes. Clearly, the square of this value would be too conservative, yielding very small step sizes and providing  substantially slower convergence. 

\subsection{Image Restoration Demonstration}

In the context of image restoration, most state-of-the-art algorithms take a patch-based approach. While the different algorithms differ in the models they enforce on the corrupted patches (or the prior they chose to consider, in the context a Bayesian formulation) the general scheme remains very much the same: overlapping patches are extracted from the degraded image, then restored more or less independently, before being merged back together by averaging. Though this provides an effective option, this \emph{locally-focused} approach is far from being optimal. As noted in several recent works (\cite{Sulam2014,Sulam2015,Romano2015}), not looking at the image as a whole causes inconsistencies between adjacent patches which often result in texture-like artifacts. A possible direction to seek for a more global outlook is, therefore, to allow for bigger patches.

We do not intended to provide a complete image restoration algorithm in this paper. Instead, we will show that benefit can indeed be found in using bigger patches in image restoration -- given an algorithm which can cope with the dimension increase. We present an image denoising experiment of several popular images, for increasing patch sizes. In the context of sparse representations, an image restoration task can be formulated as a Maximum a Posteriori formulation \cite{Elad2006}. In the case of a sparse dictionary, this problem can be posed as:
\begin{equation}
\min_{\z,\x_i,\A} = \frac{\lambda}{2} || \mathbf{z} - \mathbf{y}||^2_2 + \sum_i || \Fi\A\x_i - \mathbf{P}_i \mathbf{z} ||^2_2  +  \mu_i ||\x_i||_0,
\label{Eq:denoising}
\end{equation}
where $\z$ is the image estimate given the noisy observation $\y$, $\P_i$ is an operator that extracts the $i^{th}$ patch from a given image and $\x_i$ is the sparse representation of the $i^{th}$ patch. We can minimize this problem by taking a similar approach to that of the dictionary learning problem: use a block-coordinate descent by fixing the unknown image $\z$, and minimizing w.r.t the sparse vectors $\x_i$ and the dictionary (by any dictionary learning algorithm). We then fix the sparse vectors and update the image $\z$. Note that even though this process should be iterated (as effectively shown in \cite{Sulam2015}) we stick to the first iteration of this process to make a fair comparison with the K-SVD based algorithms.

\begin{figure}
\centering
\includegraphics[trim = 20 15 20 30, width = .45\textwidth]{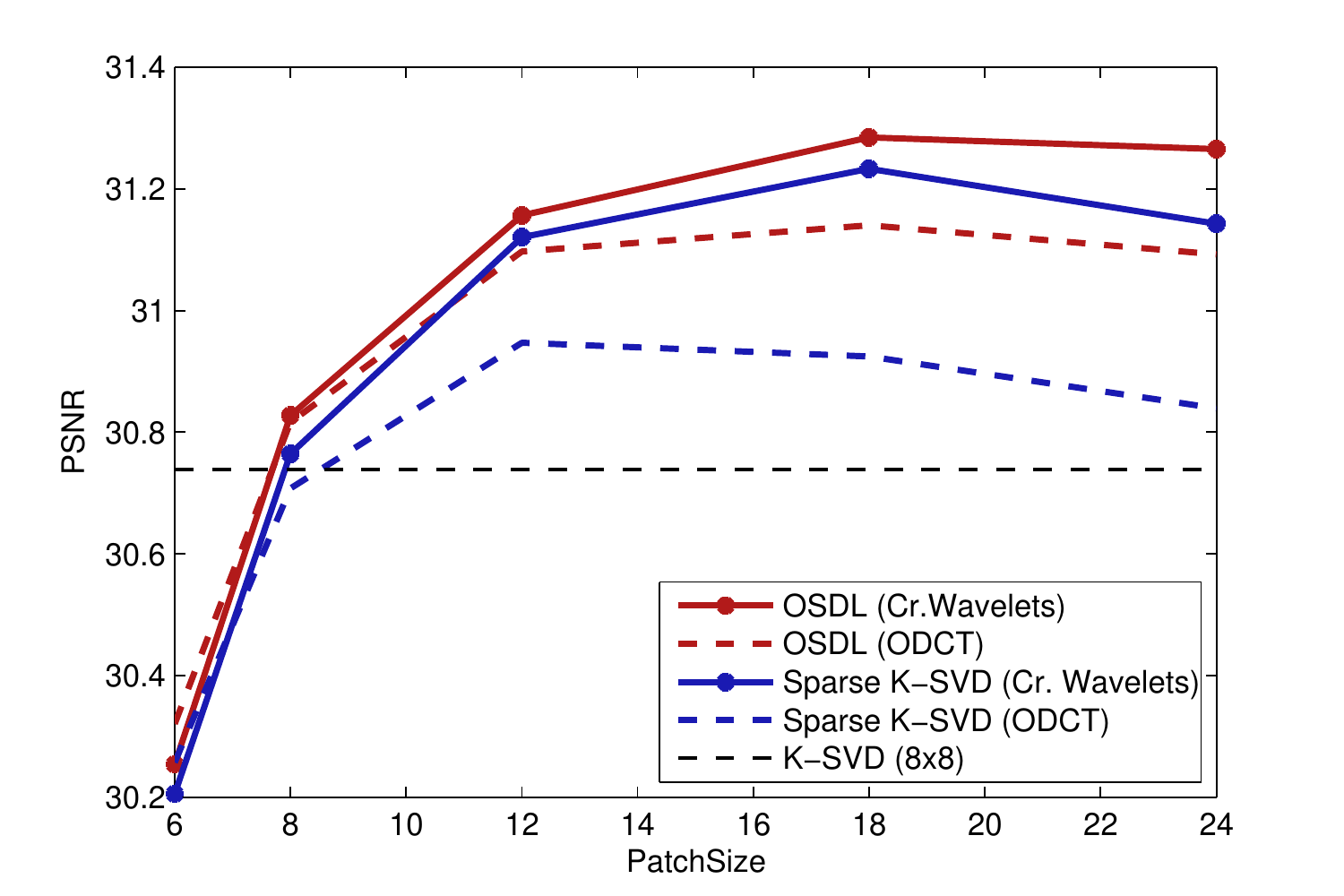}
\caption{Experiment 4: Denoising results as a function of the patch size for Sparse K-SVD and OSDL, which an overcomplete DCT dictionary and a separable cropped Wavelets dictionary.
}
\label{fig:ResultsDenoising}
\end{figure}

For this experiment, denoted as Experiment 4, we use both Sparse K-SVD and OSDL, for training the double sparsity model. Each method is run with the traditional ODCT and with the cropped Wavelets dictionary, presented in this paper. We include as a reference the results of the K-SVD denoising algorithm \cite{Elad2006}, which trains a regular (dense) dictionary with patches of size $8\times 8$. The dictionary sparsity was set to be $10\%$ of the signal dimension. Regarding the size of the dictionary, the redundancy was determined by the redundancy of the cropped Wavelets (as explained in Section \ref{Sec:CropedWave}), and setting the sparse matrix $\A$ to be square. This selection of parameters is certainly not optimal. For example, we could have set the redundancy as an increasing function of the signal dimension. However, learning such increasingly redundant dictionaries is limited by the finite data of each image. Therefore, we use a square matrix $\A$ for all patch sizes, leaving the study of other alternatives for future work.  10 iterations were used for the K-SVD methods and 5 iterations for the OSDL.

Fig. \ref{fig:ResultsDenoising} presents the averaged results over the set of 10 publicly available images used by \cite{Lebrun2013b}, where the noise standard deviation was set to $\sigma = 30$. Note how the original algorithm presented in \cite{Rubinstein2010}, Sparse K-SVD with the ODCT as the base dictionary, does not scale well with the increasing patch size. In fact, once the base dictionary is replaced by the cropped Wavelets dictionary, the same algorithm shows a jump in performance of nearly 0.4 dB. A similar effect is observed for the OSDL algorithm, where the cropped Wavelets dictionary performs the best.

\begin{figure*}
\centering
\begin{tabular}{ccc}
 \raisebox{0.1\height}{ \includegraphics[width = .3\textwidth]{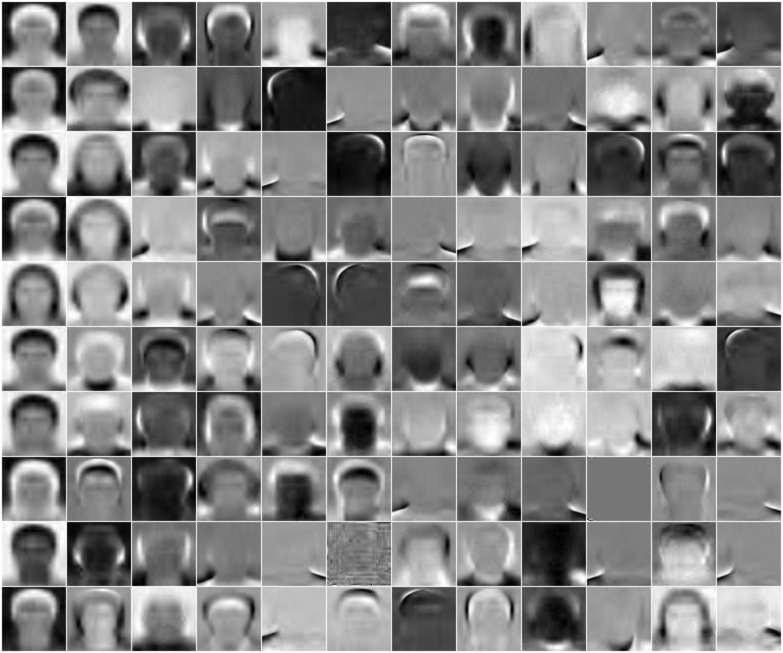}} &
 \includegraphics[trim = 20 10 20 25, width = .32\textwidth]{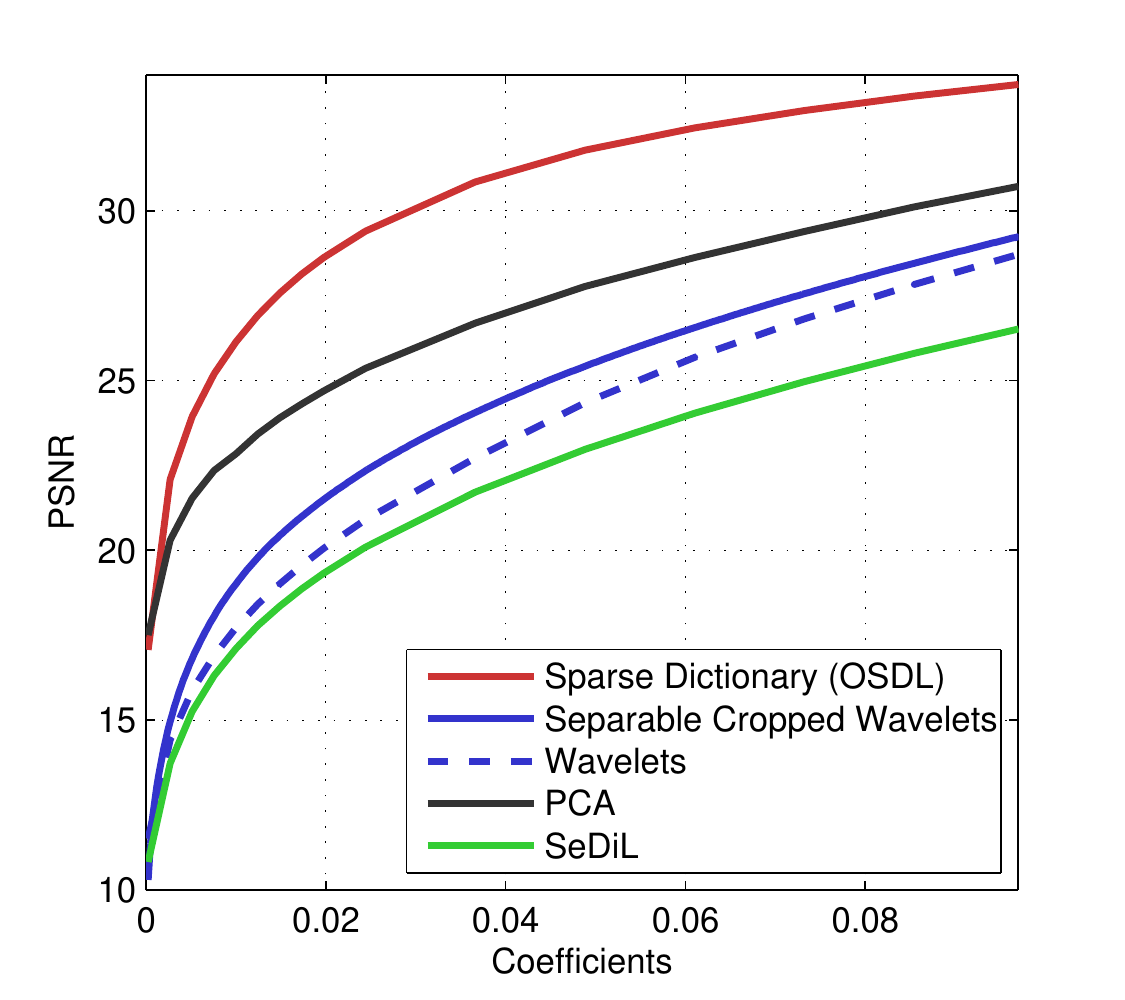} &
\includegraphics[trim = 20 10 20 25, width = .32\textwidth]{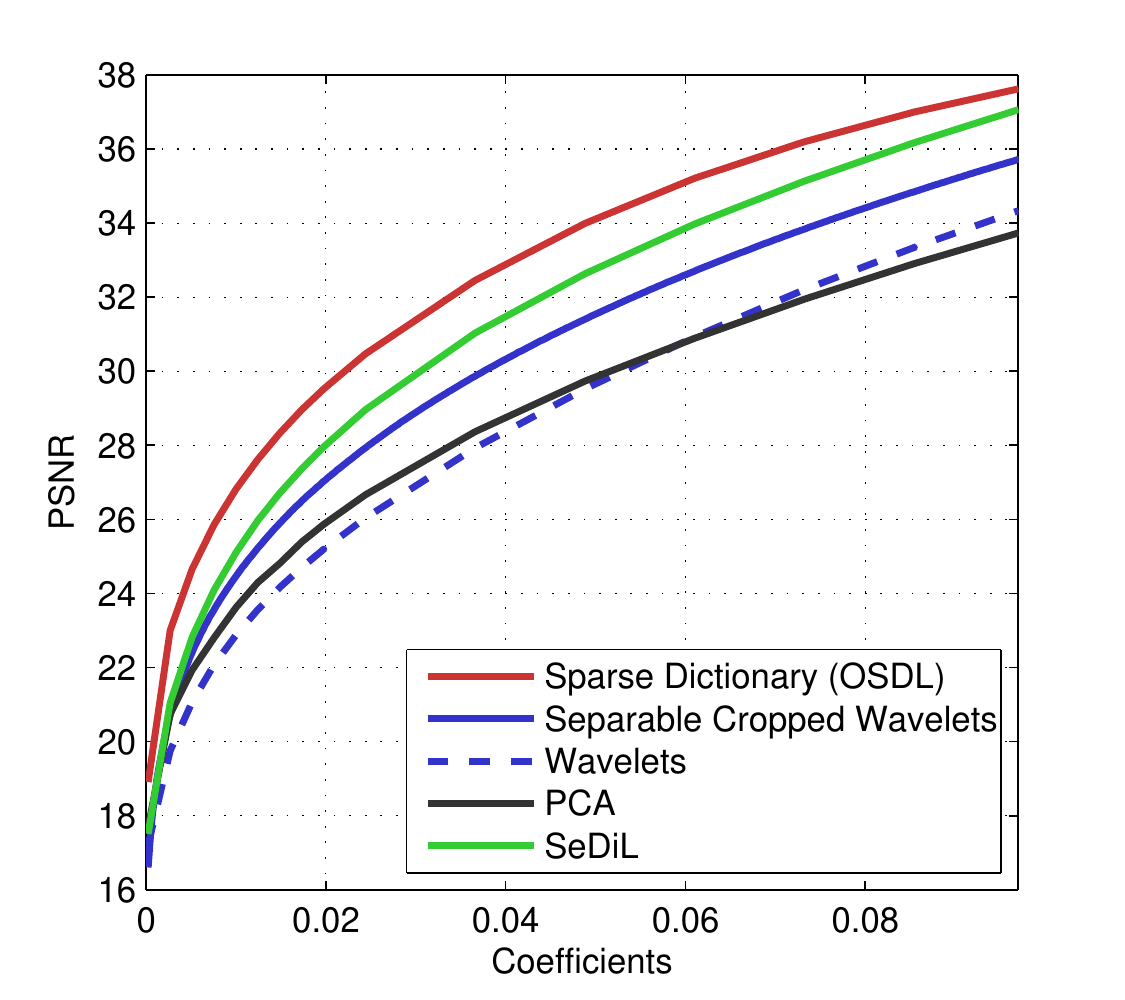} \\
$(a)$ & $(b)$ & $(c)$
\end{tabular}
\caption{Experiment 5: a) Subset of atoms from a sparse dictionary trained with OSDL on a database of aligned face images. b) Compression results (as in ratio of kept coefficients) by Wavelets, Cropped separable Wavelets, PCA, OSDL and SeDiL \cite{Hawe2013} on aligned faces. c) Compression results for the ``Cropped Labeled Faces in the Wild'' database.}
\label{fig:Compression}
\end{figure*}

Employing even greater patch sizes eventually results in decreasing denoising quality, even for the OSDL with Cropped Wavelets. Partially, this could be caused by a limitation of the sparse model in representing fine details as the dimension of the signal grows. Also, the amount of training data is limited by the size of the image, having approximately 250,000 examples to train on. Once the dimension of the patches increases, the amount of training data might become a limiting factor in the denoising performance.

As a final word about this experiment, we note that treating all patches the same way (with the same patch size) is clearly not optimal. A multi-size patch approach has already been suggested in \cite{Levin2012}, though in the context of the Non-Local Means algorithm. The OSDL algorithm may be the right tool to bring multi-size patch processing to sparse representation-based algorithms, and this remains a topic of future work.

\subsection{Adaptive Image Compression}

Image compression is the task of reducing the amount of information needed to represent an image, such that it can be stored or transmitted efficiently. In a world where image resolution increases at a surprising rate, more efficient compression algorithms are always in demand. In this section, we do not attempt to provide a complete solution to this problem but rather show how our online sparse dictionaries approach could indeed aid a compression scheme.

Most (if not all) compression methods rely on sparsifying transforms. In particular, JPEG2000, one of the best performing and popular algorithms available, is based on the 2-D Wavelet transform.
Dictionary learning has already been shown to be beneficial in this application. In \cite{Bryt2008}, the authors trained several dictionaries for patches of size $15\times 15$ on pre-aligned face pictures. These offline trained dictionaries were later used to compress images of the same type, by sparse coding the respective patches of each picture. The results reported in \cite{Bryt2008} surpass those by JPEG2000, showing the great potential of similar schemes.

In the experiment we are presenting here (Experiment 5), we go beyond the locally based compression scheme and propose to perform naive compression by just keeping a certain number of coefficients through sparse coding, where each signal is the entire target image. To this end, we use the same data set as in \cite{Bryt2008} consisting of over 11,000 examples, and re-scaled them to a size of $64\times 64$. We then train a sparse dictionary on these signals with OSDL, using the cropped Wavelets as the base dictionary for 15 iterations. For a fair comparison with other non-redundant dictionaries, in this case we chose the matrix $\A$ such that the final dictionary is non-redundant (a rectangular tall matrix). A word of caution should be said regarding the relatively small training data set. Even though we are training just over 4000 atoms on only ~11,000 samples, these atoms are \emph{only} 250-sparse. This provides a great reduction to the degrees of freedom during training. A subset of the obtained atoms can be seen in Fig. \ref{fig:Compression}a.

For completion, we include here the results obtained by the SeDiL algorithm \cite{Hawe2013} (with the code provided by the authors and with the suggested parameters), which trains a separable dictionary consisting of 2 small dictionaries of size $64\times 128$. Note that this implies a final dictionary which has a redundancy of 4, though the degrees of freedom are of course limited due to the separability imposed.

The results of this naive compression scheme are shown in Fig. \ref{fig:Compression}b for a testing set (not included in the training). As we see, the obtained dictionary performs substantially better than Wavelets -- on the order of 8 dB at a given coefficient count. Partially, the performance of our method is aided by the cropped Wavelets, which in themselves perform better than the regular 2-D Wavelet transform. However, the adaptability of the matrix $\A$ results in a much better compression-ratio. A substantial difference in performance is obtained after training with OSDL, even while the redundancy of the obtained dictionary is less (by about half) than the redundancy of its base-dictionary. The dictionary obtained by the SeDiL algorithm, on the other hand, has difficulties learning a completely separable dictionary for this dataset, in which the faces, despite being aligned, are difficult to approximate through separable atoms.

As one could observe from the obtained dictionary atoms by our method, some of them might resemble PCA-like basis elements. Therefore we include the results by compressing the testing images with a PCA transform, obtained from the same training set -- essentially, performing a dimensionality reduction. As one can see, the PCA results are indeed better than Wavelets due to the regular structure of the aligned faces, but they are still relatively far from the results achieved by OSDL.

Lastly, we show that this naive compression scheme, based on the OSDL algorithm, does not rely on the regularity of the aligned faces in the previous database. To support this claim, we perform a similar experiment on images obtained for the ``Cropped Labeled Faces in the Wild Database'' \cite{Sanderson2009}. This database includes images of subjects found on the web, and its \emph{cropped} version consists of $64\times 64$ images including only the face of the different subjects. These face images are in different positions, orientations, resolutions and illumination conditions. We trained a dictionary for this database, which consists of just over 13,000 examples, with the same parameter as in the previous case, and the compression is evaluated on a testing set not included in the training.   An analogous training process was performed with SeDiL. As shown in Fig. \ref{fig:Compression}c, the PCA results are now inferior, due to the lack of regularity of the images. The separable dictionary provided by SeDiL performs better in this dataset, whose examples consists of truncated faces rather than heads, and which can be better represented by separable atoms. Yet, its representation power is compromised by its complete separability when compared to OSDL, with a 1 dB gap between the two.

\subsection{Pursuing Universal Big Dictionaries}

Dictionary learning has shown how to truly take advantage of sparse representations in specific domains, however dictionaries can also be trained for more general domains (i.e., natural images). For relatively small dimensions, several works have demonstrated that it is possible to train general dictionaries on patches extracted from non-specific natural images. Such general-purpose dictionaries have in turn been used in many applications in image restoration, outperforming analytically-defined transforms  \cite{Bruckstein2009}.
\begin{figure*}
\centering
\includegraphics[ width = .96\textwidth]{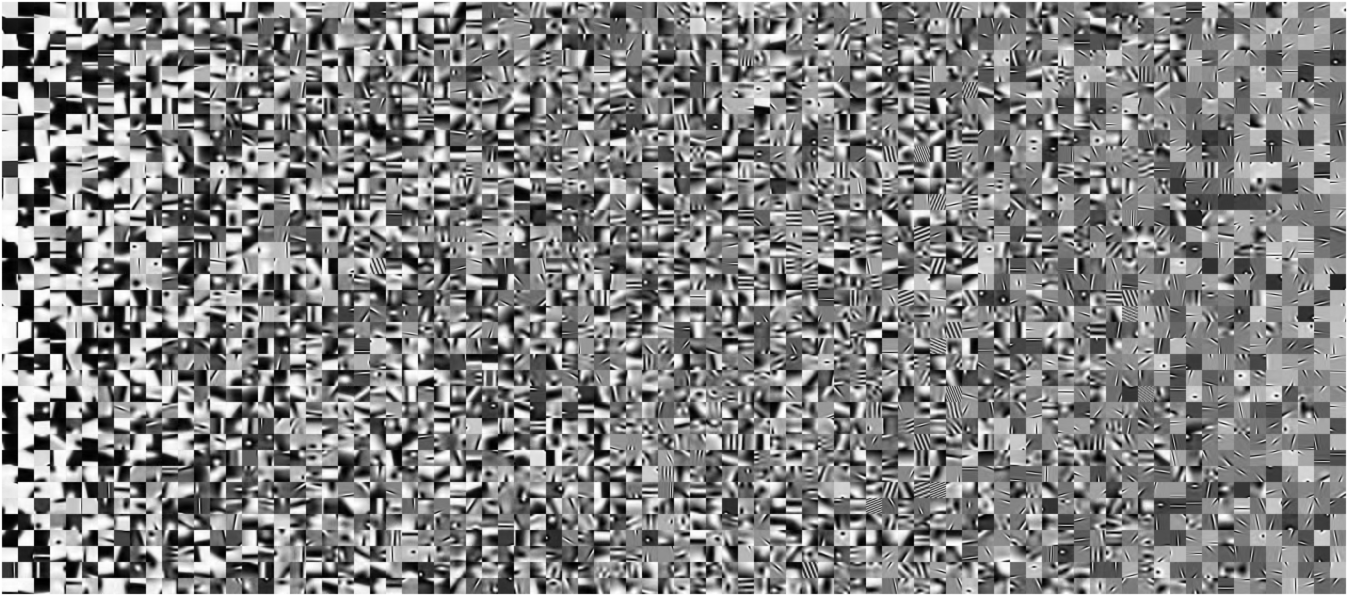}
\caption{Experiment 6: Subset of the general (sparse) dictionary for patches of size $32\times 32$ obtained with OSDL trained over 10 million patches from natural images.}
\label{fig:BigDict}
\vspace{-0.3cm}
\end{figure*}

Using our algorithm we want to tackle the training of such universal dictionaries for image patches of size $32\times 32$, i.e., of dimension 1024. To this end, in this experiment we train a sparse dictionary with a total redundancy of 6: the cropped Wavelets dictionary introduces a redundancy of around 3, and the matrix $\A$ has a redundancy of 2. The atom sparsity was set to 250, and each example was coded with 60 non-zeros in the sparse coding stage. Training was done on 10 Million patches taken from natural images from the Berkeley Segmentation Dataset \cite{MartinFTM01}. We run the OSDL algorithm for two data sweeps. For comparison, we trained a full (unconstrained) dictionary with ODL with the same redundancy, on the same database and with the same parameters.

We evaluate the quality of such a trained dictionary in an M-Term approximation experiment on 600 patches (or little images). Comparison is done with regular and separable cropped Wavelets (the last one being the base-dictionary of the double sparsity model, and as such the starting point of the training). We also want to compare our results with the approximation achieved by more sophisticated multi-scale transforms, such as Contourlets.  Contourlets are a better suited multi-scale analysis for two dimensions, providing an optimal approximation rate for piece-wise smooth functions with discontinuities along twice differentiable curves \cite{do2005contourlet}. This is a slightly redundant transform due to the Laplacian Pyramid used for the multi-scale decomposition (redundancy of 1.33). Note that, traditionally, hard-thresholding is used to obtain an M-term approximation, as implemented in the code made available by the authors. However, this is not optimal in the case of redundant dictionaries. We therefore construct an explicit Contourlet synthesis dictionary, and apply the same greedy pursuit we employ throughout the paper. Thus we fully leverage the approximation power of this transform, making the comparison fair.

Moreover, and to provide a complete picture of the different transforms, we include also the results obtained for a \emph{cropped} version of Contourlets.
Since Contourlets are not separable we use a 2-D extension of our cropping procedure detailed in Section \ref{Sec:CropedWave} to construct a cropped Contourlets synthesis dictionary.
The lack of  separability makes this dictionary considerably less efficient computationally. As in cropped Wavelets, we naturally obtain an even more redundant dictionary (redundancy factor of 5.3)\footnote{Another option to consider is to use undecimated multi-scale transforms. The Undecimated Wavelet Transform (UDWT) \cite{Mallat_AwaveletTour} and the Nonsubsampled Contourlet Transform (NSCT) \cite{NonDecimContourlets} are shift-invariant versions of the Wavelet and Contourlet transforms, respectively, and are obtained by skipping the decimation step at each scale. 
This greater flexibility in representation, however, comes at the cost of a huge redundancy, which becomes a prohibiting factor in any pursuing scheme. A similar undecimated scheme could be proposed for the corresponding cropped transforms, however, but this is out of the scope of this work.}. 

\begin{figure}
\begin{center}
\includegraphics[ width = .343\textwidth]{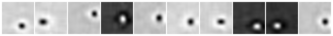}
\includegraphics[ width = .343\textwidth]{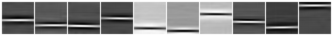}
\includegraphics[ width = .343\textwidth]{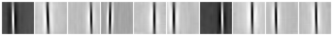}
\includegraphics[ width = .343\textwidth]{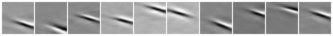}
\caption{Experiment 6: Atoms of size $32\times32$ with recurring patterns at different locations.}
\label{fig:SimilarAtoms}
\end{center}
\vspace{-0.8cm}
\end{figure}

A subset of the obtained dictionary is shown in Fig. \ref{fig:BigDict}, where the atoms have been sorted according to their entropy. Very different types of atoms can be observed: from the piece-wise-constant-like atoms, to textures at different scales and edge-like atoms. It is interesting to see that Fourier type atoms, as well as Contourlet and Gabor-like atoms, naturally arise out of the training.
In addition, such a dictionary obtains some flavor of shift invariance.
As can be seen in Fig. \ref{fig:SimilarAtoms}, similar patterns may appear in different locations in different atoms. An analogous question could be posed regarding rotation invariance.
Furthermore, we could consider enforcing these, or other, properties explicitly in the training.
These, and many more questions, are the lines of on-going work.

\begin{figure*}[t]
\centering
\begin{tabular}{ccc}
{ \includegraphics[trim = 50 10 0 5 ,width = .32\textwidth]{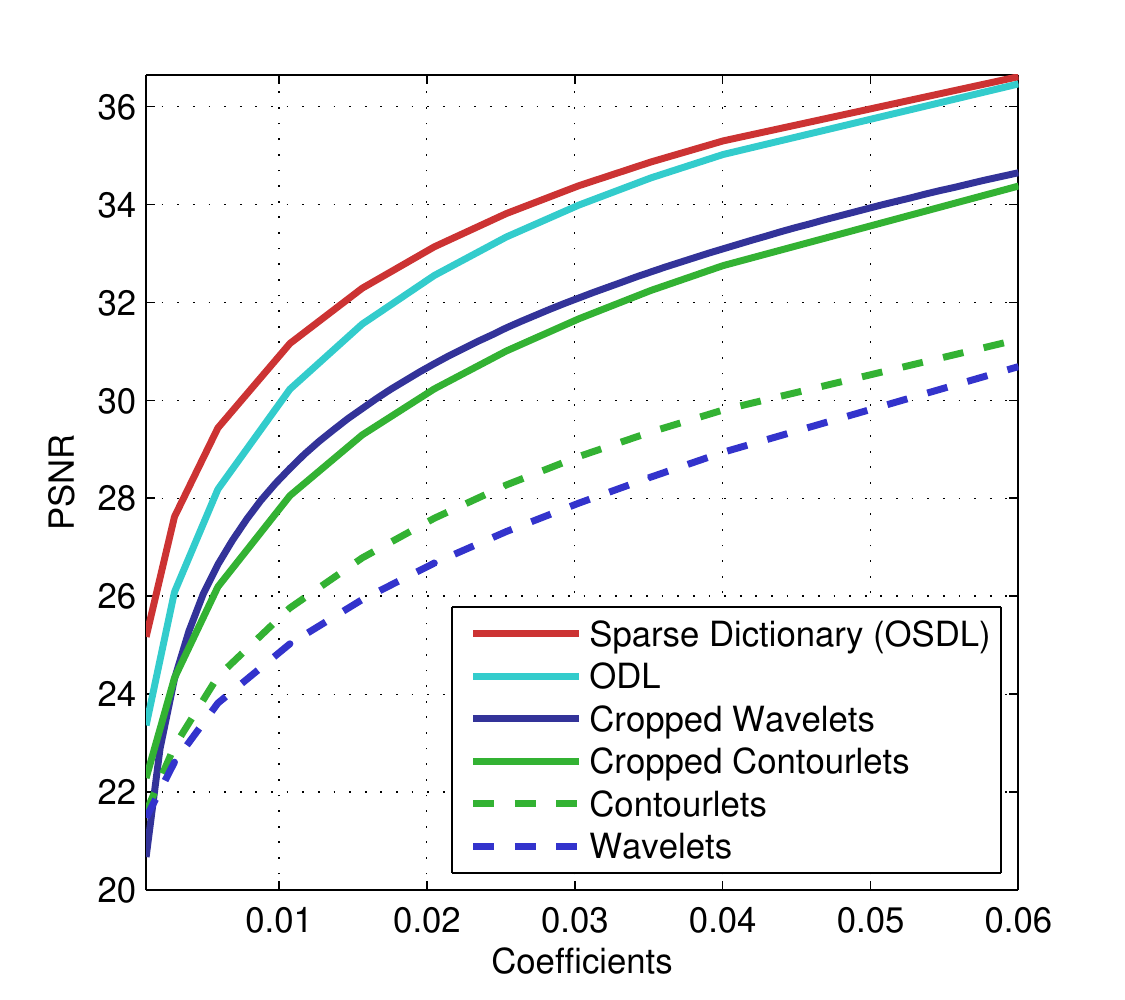}} &
 \includegraphics[trim = 40 10 10 5 ,width = .32\textwidth]{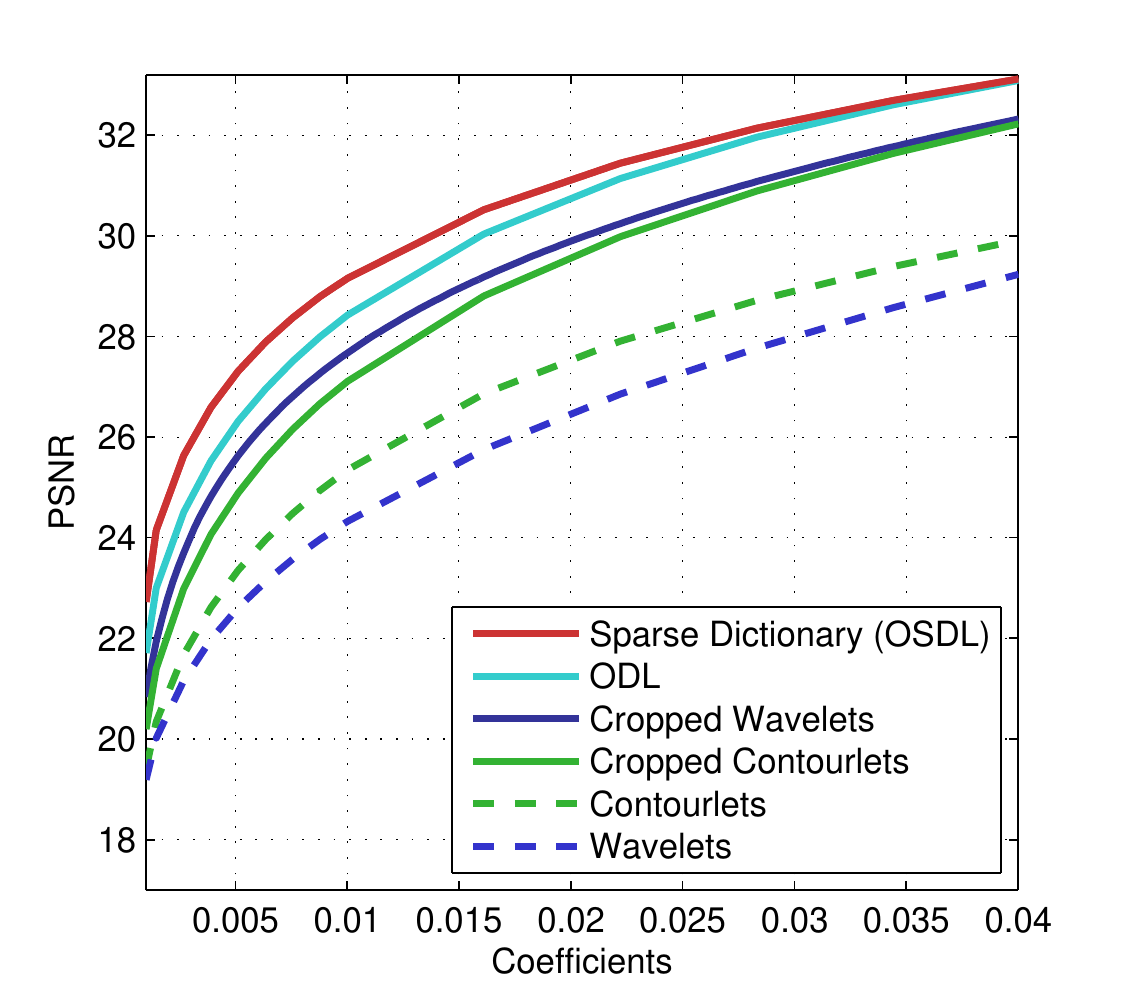} &
\raisebox{0.05\height}{\includegraphics[width = .245\textwidth]{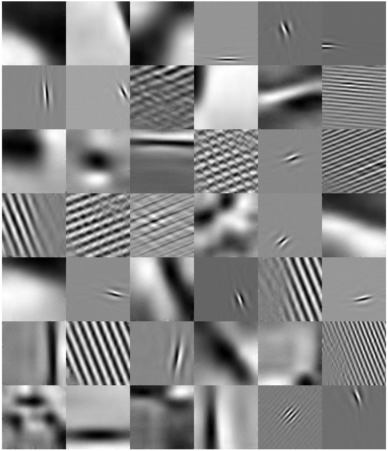}} \\
$(a)$ & $(b)$ & $(c)$
\end{tabular}
\caption{Experiment 7-8: a) M-term approximation of general image patches of size 32x32 for different methods. b) M-term approximation of general image patches of size 64x64 for different methods. c) Some atoms of size $64\times 64$ from the dictionary trained with OSDL.
\vspace{-0.4cm}}
\label{fig:Exper78}
\end{figure*}
The approximation results are shown in Fig. \ref{fig:Exper78}.a, where Contourlets can be seen to perform slightly better than Wavelets. The cropping of the atoms significantly enhances the results for both transforms, with a slight advantage for cropped Wavelets over cropped Contourlets.
The Trainlets, obtained with OSDL, give the highest PSNR. Interestingly, the ODL algorithm by \cite{Mairal2010} performs slightly worse than the proposed OSDL, despite the vast database of examples. In addition, the learning (two epochs) with ODL took roughly 4.6 days, whereas the OSDL took approximately 2 days\footnote{This experiment was run on a 64-bit operating system with an Intel Core i7
microprocessor, with 16 Gb of RAM, in Matlab.}. As we see, the sparse structure of the dictionary is not only beneficial in cases with limited training data (as in Experiment 1), but also in this big data scenario. We conjecture that this is due to the better guiding of the training process, helping to avoid local minima which an uncontrained dictionary might be prone to.

%

As a last experiment, we want to show that our scheme can be employed to train an adaptive dictionary for even higher dimensional signals. In Experiment 8, we perform a similar training with OSDL on patches (or images) of size $64\times64$, using an atom sparsity of 600. The cropped Wavelets dictionary has a redundancy of 2.44, and we set $\A$ to be square.

In order to have a fair comparison, and due to the extensive time involved in running ODL, we first ran ODL for 5 days, giving it sufficient time for convergence. During this time ODL accessed 3.8 million training examples. We then ran OSDL using the same examples\footnote{The provided code for ODL is not particularly well suited for cluster-processing (needed for this experiment), and so the times involved in this case should not be taken as an accurate run-time comparison.}.


As shown in Fig. \ref{fig:Exper78}.b, the relative performance of the different methods is similar to the previous case. Trainlets again gives the best approximation performance, giving a glimpse into the potential gains achievable when training can be effectively done at larger signal scales.
It is not possible to show here the complete trained dictionary, but we do include some selected atoms from it in Fig.  \ref{fig:Exper78}.c. We obtain many different types of atoms: from the very local curvelets-like atoms, to more global Fourier atoms, and more.

\vspace{-0.25cm}
\section{Summary and Future Work}
\label{Sect:Summary}
This work shows that dictionary learning can be up-scaled to tackle a new level of signal dimensions.
We propose a modification on the Wavelet transform by constructing two-dimensional separable \emph{cropped} Wavelets, which allow a multi-scale decomposition of patches without significant border effects.
We apply these Wavelets as a base-dictionary within the Double Sparsity model, allowing this approach to now handle larger and larger signals.
In order to handle the vast data sets needed to train such a big model, we propose an Online Sparse Dictionary Learning algorithm, employing SGD ideas in the dictionary learning task.  
We show how, using these methods, dictionary learning is no longer limited to small signals, and can now be applied to obtained Trainlets, high dimensional trainable atoms.

While OMP proved sufficient for the experiments shown in this work, considering other sparse coding algorithms might be beneficial. In addition, the entire learning algorithm was developed using a strict $l_0$ pseudo-norm, and its relaxation to other convex norms opens new possibilities in terms of training methods. Another direction is to extend our model to allow for the adaptability of the separable base-dictionary itself, incorporating ideas of separable dictionary learning thus providing a completely adaptable structure. Understanding quantitatively how different parameters affect the learned dictionaries, such as redundancy and atom sparsity, will provide a better understanding of our model. These questions, among others, are part of ongoing work.

\vspace{-0.25cm}
\section{Acknowledgements}
 The authors would like to thank the anonymous reviewers who helped improve the quality of this manuscript, as well as the authors of \cite{Hawe2013} for generously providing their code and advice for comparison purposes.

\vspace{-0.2cm}
\bibliographystyle{ieeetr}
\bibliography{MyBib}

\end{document}